\documentclass{article} % For LaTeX2e
\usepackage{iclr2027_conference,times}

\iclrfinalcopy

\usepackage{amsmath,amsfonts,bm}

\def\eqref#1{equation~\ref{#1}}
\def\1{\bm{1}}

\DeclareMathAlphabet{\mathsfit}{\encodingdefault}{\sfdefault}{m}{sl}
\SetMathAlphabet{\mathsfit}{bold}{\encodingdefault}{\sfdefault}{bx}{n}

\usepackage{hyperref}
\usepackage{url}
\usepackage[utf8]{inputenc} % allow utf-8 input
\usepackage[T1]{fontenc}    % use 8-bit T1 fonts
\usepackage{hyperref}       % hyperlinks
\usepackage{url}            % simple URL typesetting
\usepackage{booktabs}       % professional-quality tables
\usepackage{amsfonts}       % blackboard math symbols
\usepackage{nicefrac}       % compact symbols for 1/2, etc.
\usepackage{graphicx}
\usepackage{subcaption}
\usepackage[table]{xcolor}
\usepackage{colortbl}
\usepackage{array}
\usepackage{amsmath}

\definecolor{matplotlibC3}{HTML}{d62728}

\definecolor{darkgreen}{RGB}{0,90,0}

\newcolumntype{P}[1]{>{\centering\arraybackslash}p{#1}}

\title{Interpreting Brain Responses to Language\\ with Sparse Features from Language Models}

\author{%
  Michael A. Lepori \\
  Dept. of Computer Science\\
  Brown University\\
  \texttt{michael\_lepori@brown.edu} \\
  \And
   Kendrick Kay\\
   Dept. of Radiology\\
   University of Minnesota\\
   \texttt{kay@umn.edu} \\
   \And
  Greta Tuckute \\
  Kempner Institute \\
  Harvard University \\
  \texttt{gtuckute@fas.harvard.edu} \\
}

\begin{document}

\maketitle

\begin{abstract}
A central goal of cognitive neuroscience is to characterize the features that are represented by human language cortex.
Artificial language models (LMs) have emerged as a powerful tool to address this challenge, but studies relating biological and artificial representations are often criticized as relating one black box to another. 
The present work introduces \textbf{Augmented Sparse Encoding Models}, an encoding framework that replaces dense LM hidden states with hierarchically-organized sparse autoencoder (SAE) features, while explicitly including surprisal as a predictor.
Using this approach, we (i) produce \textit{interpretations} of neural responses and (ii) test whether model-brain alignment reflects primary or idiosyncratic variation in LM representations.
Using a high-field 7T fMRI dataset of eight participants listening to 200 linguistically diverse sentences, we first validate our modeling framework by recovering previous interpretations of voxel populations tuned to processing difficulty and meaning abstractness. We then interpret a previously-uncharacterized (but reliable) voxel population and find that it is tuned to people-related content. Next, we show that the fronto-temporal human language network is predicted by a common set of features across its constituent regions, but find that frontal regions are relatively well-explained by surprisal alone, even in the absence of LM-based features.
Finally, we show that %brain responses during language processing are not merely predictable from an arbitrary set of LM features. Rather, 
brain responses are best explained by the features that tend to capture the most general information encoded in LM representations, suggesting a nontrivial correspondence between brain and LM language representation.
\end{abstract}

\section{Introduction}

Humans effortlessly map speech signals to complex meanings through language, but the representations that support this process in the brain remain poorly characterized. Language models (LMs) have become a central tool for probing the representations underlying language processing \citep{jain2018incorporating,caucheteux2022brains,tuckute2024language}. At the same time, this line of work has been criticized as relating one black box (an LM) to another (the brain) without yielding clear claims about what neural populations represent or compute. 
Recently, two developments make it increasingly feasible to gain scientific insight from LM encoding models. On the LM side, sparse autoencoders (SAEs) provide a tool for decomposing dense LM hidden states into latent features that are often more identifiable and easier to interpret than residual stream features \citep{bricken2023monosemanticity}. 
On the neuroscience side, decades of neuroimaging work have delineated a
fronto-temporal language network \citep{binder1997human,fedorenko2024language}, and additional work has made progress in characterizing the response properties of these regions. One line of work shows that responses in language regions track linguistic processing difficulty (as quantified by LM surprisal, for instance; \citealp{henderson2016language,shain2020fmri,wehbe2021incremental,heilbron2022hierarchy,tuckute2024driving}), while another shows tracking of concrete vs. abstract meanings during sentence or narrative comprehension (\citealp{botch2024neural,tuckute2025two}; with related work on single-words, e.g., \citealp{binder2005distinct,west2000imaginal,fernandino2015predicting}).

In the present study, we leverage these advancements to propose and validate a new class of encoding models: \textbf{Augmented Sparse Encoding Models}.\footnote{Code available \href{https://github.com/mlepori1/Interpretable_Encoding_Models}{here}.} Compared to standard LM encoding models, our framework introduces two changes: we project dense residual-stream LM features into a sparse, hierarchically organized SAE basis, and we augment that basis with an explicit feature that captures processing difficulty (surprisal). \textbf{Our first goal is to produce interpretations of voxel response tuning.}
Next, we characterize properties of the features that drive alignment, going beyond prior LM encoding work that demonstrates that LM representation can predict neural responses. To do so, we leverage an SAE basis that is hierarchically organized from general, primary features to fine-grained, idiosyncratic features.
\textbf{Our second goal, therefore, is to identify whether brain alignment relies on primary or idiosyncratic LM feature dimensions}.

Using Augmented Sparse Encoding Models, we make the following contributions:
\begin{enumerate}
    \item SAE features can predict voxel responses to language as accurately as dense LM residual-stream features, while also providing interpretations that affirm and extend prior neuroscience findings on processing difficulty-tuned and content-tuned voxels (Section~\ref{Sec:Study1}).
    \item We discover and interpret previously uncharacterized voxel populations (Section~\ref{Sec:Study2}).
    \item We show how different regions of the human language network respond differentially to processing difficulty vs. content features across three independent datasets spanning sentence and story comprehension: frontal brain regions are generally explained well by processing difficulty, whereas temporal regions draw more on LM-derived ``content'' features. We further characterize the features that are prevalent across brain regions and individuals (Section~\ref{Sec:Study3}).
    % We further characterize the features that are prevalent across brain regions and individuals (Section~\ref{Sec:Study3}).
    \item We show that brain responses to language are predicted by primary, general features of LM representations rather than idiosyncratic ones---revealing a deeper correspondence between artificial and biological language representations than implied by measures of encoding accuracy alone (Section~\ref{Sec:Study3}).
\end{enumerate}

% This feature-level approach allows us to ask not only whether LM representations predict brain responses, but whether the same interpretable features are reused across language regions and individuals, and whether brain alignment depends on general or idiosyncratic dimensions of the LM representation.

\section{Related Work}

\textbf{Organization of the Language Network}
A set of frontal and temporal brain areas in the left hemisphere---the ``language network''---supports language understanding and production across input modalities \citep{binder1997human,deniz2019representation,lipkin2022probabilistic,hu2023precision,fedorenko2024language}.
Within this system, prior work has emphasized processing difficulty and predictability as robust drivers of univariate response magnitude during comprehension \citep{shain2020fmri,wehbe2021incremental,tuckute2024driving}. A separate line of work has studied abstract–concrete semantic organization in the brain, mostly in single-word experiments  \citep{binder2005distinct,west2000imaginal,fernandino2015predicting}; one recent study links these lines of work by showing that sentence-evoked voxel responses are jointly organized along dimensions of processing difficulty and content (specifically meaning-abstractness) \citep{tuckute2025two}.
So, although the location of the language network and its broad response properties are now well established, how linguistic information is represented at a finer grain across voxels---within and across regions, and across individuals---remains less well understood.

% relatively little work on quantifying participant-specific dimensions.. 

% While this work goes a long way toward characterizing \textit{where} and \textit{selectivity }linguistic processing is implemented, it is still largely unknown how the language network itself is organized. Indeed, a large body of recent work suggests that the various areas comprising the language network behave somewhat monolithically --- producing similar response profiles that are tuned to both structural and semantic features.

\textbf{Mechanistic Interpretability} Efforts to interpret the algorithms and representations learned by trained neural networks have coalesced into a set of techniques and paradigms that comprise the field of mechanistic interpretability \citep{elhage2021mathematical, geiger2025causal}. In particular, the rediscovery of sparse dictionary learning \citep{olshausen1996emergence} has driven progress on unsupervised methods for interpreting and controlling model behavior \citep{fel2023craft, ameisen2025circuit, bricken2023monosemanticity}.  Sparse dictionary learning methods, such as SAEs, attempt to learn overcomplete dictionaries of features, such that dense representations can be approximated by a small number of monosemantic (and therefore interpretable) features. %We employ pretrained SAEs to create feature spaces for Augmented Sparse Encoding Models.

\textbf{LM Encoding Models}
Standard LM encoding studies have established that LM representations can predict brain responses to language at the granularity of fMRI voxels, M(EEG) sensors, and intracranial recordings \citep{jain2018incorporating,caucheteux2022brains,hosseini2024universality,alkhamissi2025language}. 
% Caucheteux & King, Antonello & Huth, Hosseini, ARN review (preprint from Europe folks too?), AlKhamissi,
% of course, also acknowledging earlier interpretable feature spaces, not LLMs: Wehbe 2014; Mitchell 2008, Anderson , not performant or LLM
A growing body of work has aimed to distill and interpret the properties of LM representations that enable LMs to capture neural responses to language. One direction restricts or perturbs the linguistic information (e.g., syntactic vs. semantic information) a model is trained on or can use during inference \citep{pasquiou2023information,merlin-toneva-2024-language,kauf2024lexical}. A second direction isolates specific internal LM components, such as attention weights rather than standard residual-stream embeddings \citep{lamarre2022attention,kumar2024shared}. A third direction develops explicitly human-readable or sparse, identifiable feature spaces for voxel-wise encoding models \citep{benara2024crafting,singh2025evaluating,zeng2025disentangling}. Our work is closest to the third direction. Most closely related, \citet{zeng2025disentangling} apply sparse dictionary learning to \textit{word-level} embeddings in narratives. We differ by (i) using SAEs over contextual LM hidden states, and (ii) leveraging the hierarchical structure of one class of SAE architecture to ask \textit{which levels of the LM feature hierarchy} best align with brain responses.
Concurrent and complementary work employs SAEs to study semantic organization in the cortex \citep{guo2026sparse} and to dissociate temporal windows and broad cortical regions using intracranial measurements \citep{kleinman2026back}. 

% We differ by (i) using sparse autoencoders (SAEs) over contextualized LM hidden states rather than over static word embeddings, and (ii) leveraging the hierarchical structure of one class of SAE architectures (Matryoshka SAEs) to ask \textit{which levels of the LM feature hierarchy} best align with brain responses during \textit{sentence-level} language processing.

% developing explicitly human-interpretable or sparse feature spaces for voxelwise encodings:
% Crafting questions: Benara 2024 and Singh 2025
% Zeng & Gallant (2025) sparse concept basis (over word embeddings) for interpretable voxelwise encoding in story-listening fMRI but not a pretrained hierarchical SAE over LM hidden states

\section{Methods}

\textbf{Brain Data}\label{methods:brain}
We used brain responses from prior work \citep{tuckute2025two}, in which eight proficient English speakers underwent ultra-high-field 7 Tesla (7T) fMRI while listening to 200 linguistically diverse sentences, each repeated three times in pseudorandomized order. The fMRI blood-oxygenation-level-dependent (BOLD) response amplitude was modeled with a General Linear Model \citep{glmsingle}, yielding a single beta value per voxel per sentence trial (relative to a fixation baseline). We averaged over the three repetitions of a sentence. We restricted analyses to voxels with a noise ceiling signal-to-noise ratio $>$ 0.4 (NCSNR; computed from stimulus repetitions as in \citealt{allen2022massive}), ensuring that voxel selection and subsequent interpretation are based on voxels with reliable responses. Given the left-lateralization of language \citep{lipkin2022probabilistic}, all analyses are restricted to the left hemisphere. Analyses were conducted on the fsaverage surface, but for simplicity we refer to surface vertices as ``voxels'' throughout the paper.

We select voxels of interest in two main ways: one based on two principal components (PCs) of sentence-evoked responses identified by \citet{tuckute2025two}, and the other based on whether a voxel is part of the fronto-temporal language network \citep{fedorenko2010new}.
For the PC-based selection, we leverage a recent finding that two PCs capture most generalizable variance in sentence-evoked voxel responses (\citealp{tuckute2025two}; and related work \citealp{shain2020fmri,wehbe2021incremental,botch2024neural}): PC1 reflects \textit{processing difficulty} (characterized by frequency and surprisal) and PC2 reflects concrete--abstract \textit{content} (the degree to which a sentence's meaning is grounded in perceptual experience). Based on each voxel's correlation with these two PCs, we define five populations: \textbf{Hard-to-Process} (high PC1), \textbf{Easy-to-Process} (low PC1), \textbf{Abstract} (high PC2), \textbf{Concrete} (low PC2), and \textbf{Ghost} (weak loading on both). For the language-network selection, we use five canonical left-hemisphere functional regions of interest (fROIs) defined via a \textit{sentences}~$>$~\textit{nonwords} contrast \citep{fedorenko2010new}, selecting the top $10\%$ significant voxels within three frontal brain parcels---inferior frontal gyrus (IFG), its orbital portion (IFGorb), and middle frontal gyrus (MFG)---and two temporal parcels---anterior temporal (AntTemp) and posterior temporal (PostTemp) (see Appendix~\ref{app:voxel_selection} for details).
This dataset---combining 7T resolution with highly reliable sentence-level measurements---provides an ideal testbed for the voxel-level analyses we pursue below. We additionally use two independent 3T datasets for replication, see Appendices~\ref{app:3T} and~\ref{app:rep_natstories}. 

\begin{figure}
    \centering
    \includegraphics[width=\linewidth]{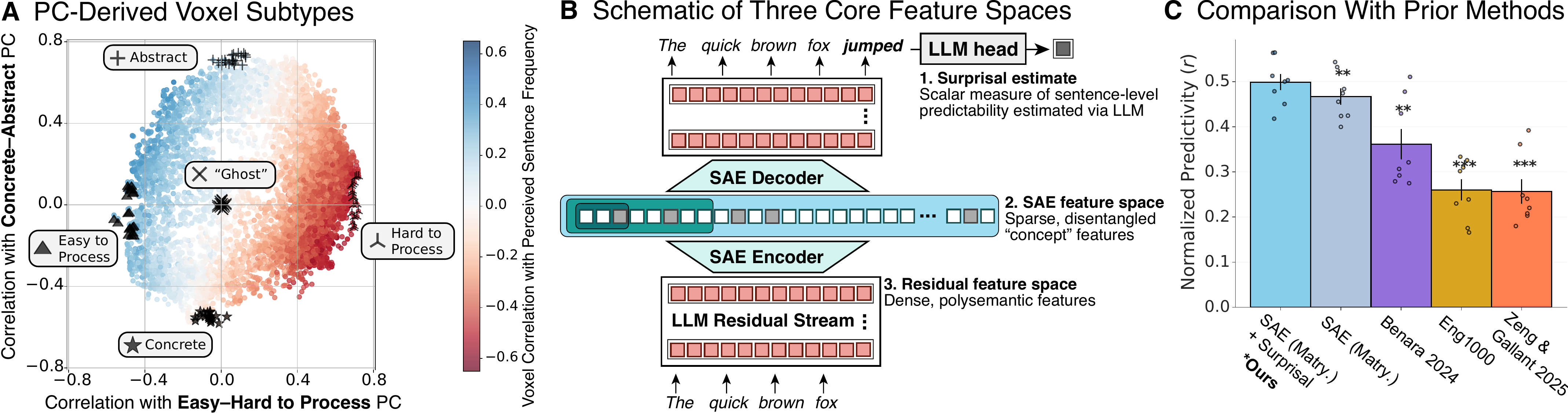}
    \caption{Overview of methods. \textbf{(A)} Each dot is a voxel from one sample participant, projected onto two PCs from prior work. Voxels selected for analysis in Study 1 and 2 are annotated. Color denotes each voxel's correlation with sentence-frequency annotations derived from humans \citep{tuckute2025two}. \textbf{(B)} Visualization of LM feature spaces. Prior LM encoding models rely on residual stream features, whereas Augmented Sparse Encoding Models use SAE feature spaces along with surprisal, which is measured using the mean of LM output probabilities over tokens in a sentence. \textbf{(C)} Comparison with prior interpretable encoding methods, evaluated on the larger set of language-fROI voxels analyzed in Section~\ref{Sec:Study3} (* = $p<.05$, ** = $p<.01$, *** = $p<.001$; see Appendix~\ref{app:analysis_details} for methods). }
    \label{fig:brain_and_features}
\end{figure}

\textbf{Augmented Sparse Encoding Models}
\label{sec:Encoding}
Encoding models are defined by two components: (i) a feature extractor that transforms language stimuli into a feature vector, and (ii) a linear readout that maps that feature vector onto neural responses to the same linguistic stimuli. LM encoding models typically use pretrained LMs as feature extractors, aggregating the intermediate residual-stream representations of an LM to create one feature vector per sentence (See Fig.~\ref{fig:brain_and_features}B, Bottom). Linear readouts are typically regularized linear regressions. See individual study sections for details on the linear readouts that we employ.

In this work, we introduce two changes to the typical framework to make the resulting LM encoding model more interpretable. First, we employ pretrained SAEs to transform dense, difficult-to-interpret residual-stream LM representations into a much larger, sparser, and (ideally) more interpretable feature basis. Specifically, we pass LM hidden states (i.e., residual stream representations) into the encoder of an SAE and compute the mean representation over tokens in a sentence in this sparse basis (Fig.~\ref{fig:brain_and_features}B, Middle).
Our second change is the inclusion of a dedicated feature that encodes the average surprisal (i.e., negative log probability) of a sentence, as computed by an LM (See Fig.~\ref{fig:brain_and_features}B, Top). Surprisal is widely used as a measure of sentence processing difficulty, and has been shown to correlate with behavioral and neural signatures of processing difficulty in human participants \citep{wilcox2020predictive, michaelov2024strong}. Together, these two changes enable us to (i) distinguish whether neural responses are best explained by processing difficulty vs. representational content obtained from the LM/SAE, and (ii) attempt to directly interpret the SAE features that explain neural responses. 

We instantiate these changes using the \texttt{gemma-2-2b} base model \citep{team2024gemma}, owing to its competitive language performance, relatively small size, and availability of pretrained SAEs. Following prior work on SAEs in gemma models, we primarily use representations from layer 12, though we replicate some of our findings with layer 14 (Appendix~\ref{app:layer14replication}). 
We analyze two different SAE architectures: JumpReLU and Matryoshka. The JumpReLU SAE is a common, high-performing variation of a standard SAE \citep{rajamanoharan2024jumping, lieberum2024gemma}. The Matryoshka SAE introduces additional structure into the SAE basis to learn more disentangled features at varying levels of granularity \citep{bussmannlearning, matryoshka_release}. Specifically, it is trained in a hierarchical fashion that produces five nested feature sets: the first set can coarsely approximate LM residual states on its own, and each subsequent set is trained to reduce the residual reconstruction error given all earlier sets. As a result, early SAE features capture general, widely-applicable information, and later features capture progressively finer-grained features.
We use pretrained JumpReLU SAEs from the \texttt{Gemma Scope} release \citep{lieberum2024gemma}, and pretrained Matryoshka SAEs from \citet{matryoshka_release}. Standard \texttt{gemma-2-2b} residual-stream embeddings have 2304 dimensions, the JumpReLU SAE has 16.4K dimensions, and the Matryoshka SAE has 32K dimensions.

% Specifically, this SAE is trained in a hierarchical fashion that results in five distinct sets of features: the first set of features can coarsely approximate LM hidden states on their own, with successive feature sets are trained to reconstruct LM hidden states in conjunction with all earlier sets. Intuitively, each successive feature set should reduce the residual reconstruction error. This approach results in feature sets that are increasingly fine-grained, with the first several dimensions capturing general, high-frequency features and the last several dimensions capturing extremely granular features. 

%%%%%%%%%%%%%%%%%%%%%%%%%%%%%%%%%%%%%%%%%%%%% STUDY 1 %%%%%%%%%%%%%%%%%%%%%%%%%%%%%%%%%%%%%%%%%%%%%

\section{Study 1: Validating Augmented Sparse Encoding Models Using PC-Derived Voxel Subtypes}
\label{Sec:Study1}

We first evaluate our Augmented Sparse Encoding Models on the four PC-derived voxel subtypes, defined using two brain-derived PCs \citep{tuckute2025two}: one corresponding to processing difficulty (easy-to-process to hard-to-process) and another corresponding to content (concrete to abstract sentence meanings) (Fig.~\ref{fig:brain_and_features}A, Section~\ref{methods:brain}). These PC interpretations were obtained via a small set of manually collected (human-derived) annotations. Thus, these voxel populations provide a useful validation setting: because their response properties have already been partially characterized, we can ask whether \textit{interpretations} obtained from Augmented Sparse Encoding Models recover or contradict these prior accounts. 
%Thus, this validation requires more than just matching predictive performance.

\textbf{Analysis Methods}
We wish to identify a sparse set of features that best predict a voxel's response profile in order to gain insight into the features driving that voxel's activity. To do so, we employ a two-stage analysis pipeline comprised of feature selection (using a LASSO regression) and then refitting (using a Ridge regression). We always include the surprisal feature in the Ridge regression, allowing for clean comparisons between full LM encoding model predictivity and a surprisal-only baseline. This pipeline is fully cross-validated (5-fold). The final predictivity score is defined as the mean Fisher-$z$-transformed correlation across folds, noise-ceiling normalized at the voxel level.
See Appendix~\ref{app:analysis_details} for more details.

\begin{figure}
    \centering
    \includegraphics[width=\linewidth]{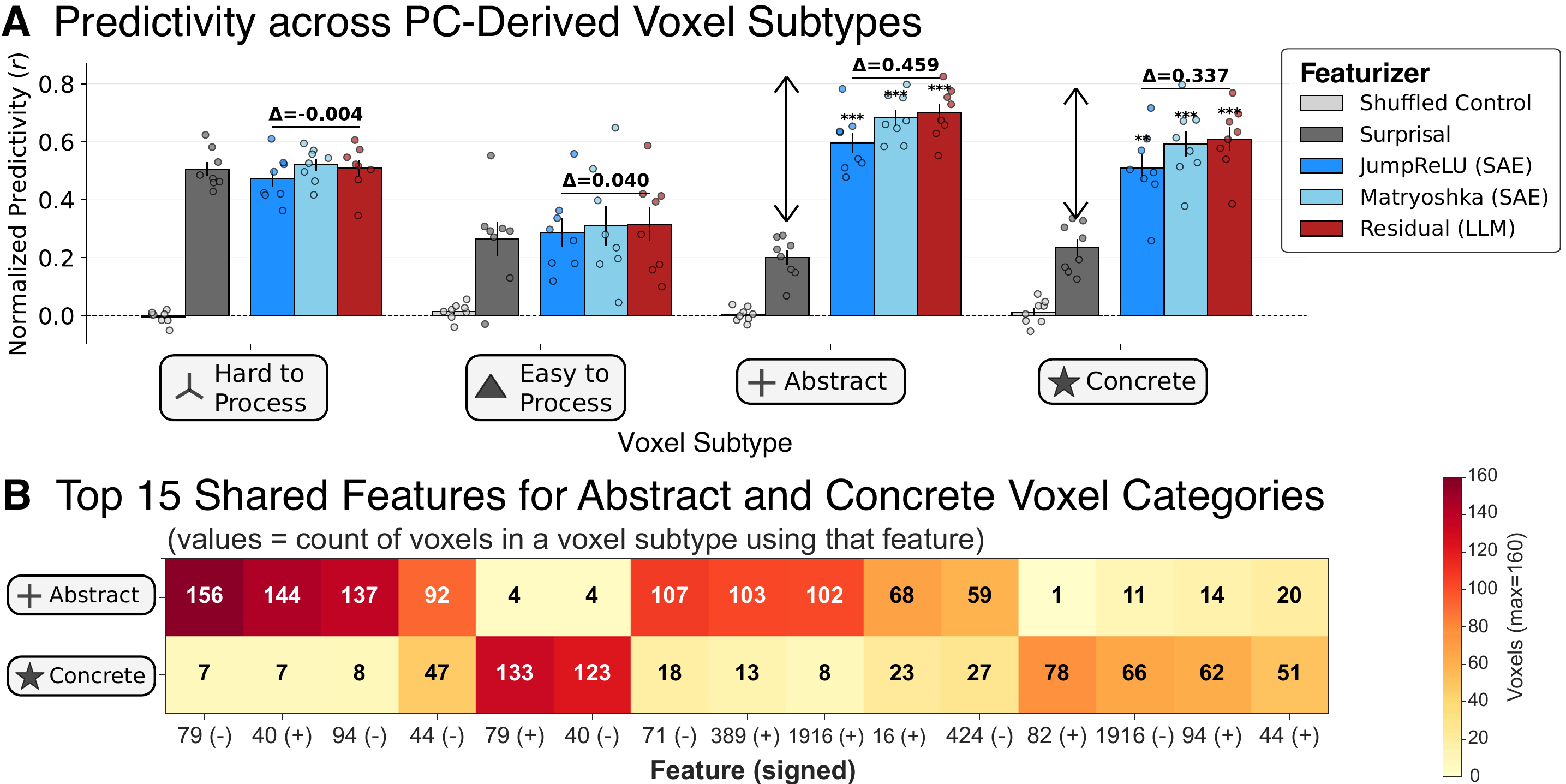}
    \caption{\textbf{(A)} Normalized encoding model predictivity of four voxel subtypes using different feature spaces. Processing-difficulty voxels show no predictivity benefit from LM representation features beyond surprisal, whereas abstract and concrete voxels benefit substantially. $\Delta$ is the difference in mean predictivity of surprisal alone versus the mean of the three featurizers (JumpReLU, Matryoshka, Residual). \textbf{(B)} Matryoshka SAE features selected by regressions on \textbf{Abstract} and \textbf{Concrete} voxels. Features form different subspaces for each subtype generalize across individual brains, and correspond to features that match---and extend---existing interpretations of these voxel populations.}
    \label{fig:voxcat_pred}
\end{figure}

\textbf{Finding 1: Sparse SAE features predict voxel responses as well as dense residual-stream features.}
From Fig.~\ref{fig:voxcat_pred}A, we see that SAE features tend to predict voxel responses to language stimuli nearly as well as dense LM residual features. In particular, features in the Matryoshka SAE basis achieve very similar performance as features in the dense residual LM basis, across all four voxel subtypes (on average, 18 Matryoshka SAE features were selected per voxel; Appendix~\ref{app:support_size}). The JumpReLU basis consistently yields feature sets that are somewhat worse for prediction. All feature spaces outperform a control that attempts to predict voxel responses from mismatched LM features \citep{hadidi2026spurious}. 
Compared to prior methods, our Augmented Sparse Encoding Models outperform several prior interpretable feature spaces \citep{benara2024crafting,huth2016natural,zeng2025disentangling} (Fig.~\ref{fig:brain_and_features}C; See Appendix~\ref{app:analysis_details} for details).

%Overall, this analysis shows that sparse Matryoshka SAE features can reliably predict voxel responses to language, while also allowing for feature interpretation.

\textbf{Finding 2: Dissociation between processing difficulty-driven and content-driven voxel regimes.}
Fig. \ref{fig:voxcat_pred}A shows a clear dissociation across the PC-derived subtypes: voxels in the \textbf{Hard-to-Process} subtype are captured by surprisal alone, with no significant improvement from SAE or residual-stream features. In contrast, voxels in the \textbf{Abstract} and \textbf{Concrete} subtypes benefit significantly from the inclusion of SAE or residual-stream features, above and beyond surprisal (Fig.~\ref{fig:voxcat_pred}A, $p<0.001$). This finding recovers---at the single-voxel level---a dissociation between processing difficulty-driven and content-driven voxel populations during language comprehension. Consistent with prior work that summarized voxel responses via PCA or region-level averaging (\citealp{tuckute2025two}, also e.g., \citealp{wehbe2021incremental}), we show this here using  Augmented Sparse Encoding Models.

\textbf{Finding 3: Distinct features refine interpretations of content-driven voxels.}
We next ask which specific model features support predictivity of the content-driven voxel subtypes. SAE features can often (but not always) be interpreted by examining the natural-language contexts that maximally activate them, using resources such as Neuronpedia \citep{neuronpedia}. 
For this analysis, we focus on the Matryoshka SAE, as it (i) better matched the predictivity of the residual-stream feature space, (ii) is known to produce more interpretable features than JumpReLU SAEs \citep{bussmannlearning}, and (iii) results in more sparse feature bases than the JumpReLU SAE (Appendix~\ref{app:support_size}). For each voxel, we refit the encoding model on all sentences and examine the (signed) SAE features selected by the model.
Because this feature basis is learned without supervision and contains tens of thousands of candidate features, agreement with prior interpretations would be nontrivial.

\begin{table}[t]
\centering
\rowcolors{2}{gray!12}{white} % start at row 2 (keeps header unshaded)
\begin{tabular}{p{1.25cm} p{1.75cm} p{3cm}p{6.5cm}}

\hline
\rowcolor{white}\textbf{Feat. \#} & \textbf{Feat. Label} & \textbf{Annotation} & \textbf{Max-Activating Examples} \\
\hline

94 & People & Descriptions of public figures involved in events & \parbox[t]{6.5cm}{%
\textit{- She worked under execute chef Walter Scheib III, who resigned...}\\
\textit{- Singer and former Haitian presidential candidate, picture here...}\\
}\\

79 & Scenery & Descriptions of locations and scenes & \parbox[t]{6.5cm}{%
\textit{- lost control and crashed into a ditch}\\
\textit{- apartment complex's newly created dog park}\\
}\\

40 & Emotions & Informal speech related to emotions and relationships & \parbox[t]{6.5cm}{%
\textit{- oh shit cause I still have mad feelings for this girl right?}\\
\textit{- Yeah I'm bummed. To the max.}\\
}\\

% 44 & Short Sentences & Greater activation on sentences with few tokens. & *This feature activates strongly on the first token. We take the mean of SAE activations, so this tracks the number of tokens in the sentence.* \\

49 & Pronouns &  Possessive pronouns. & \textit{his}, \textit{their}, \textit{our}, \textit{my} \\

44, 71 & Token Count & Greater activation on sentences with few tokens. & *This feature activates on the first token. We take the mean of SAE activations, so this tracks the number of tokens in a sentence.* \\

389 & Questions & Questions about ideas, processes, or existence. & \parbox[t]{6.5cm}{%
\textit{- Does anyone have a more precise idea?}\\
\textit{- Does one exist?}\\
}\\

1916 & Consequences & Statements about consequences or results of processes. & \parbox[t]{6.5cm}{%
\textit{- from the initiation to the completetion of...}\\
\textit{- Consequently, it does not appear that the decline...}\\
}\\

16 & Failure & Statements about destruction, failure, or error. & \parbox[t]{6.5cm}{%
\textit{- a slow-moving train-wreck that would do economic damage...}\\
\textit{- it will always crash halfway through...}\\
}\\

\hline
\end{tabular}

\caption{Descriptions of Matryoshka SAE features that often predict voxel responses. Feature \# corresponds to the dimension in the Matryoshka latent space. Feature Label and Annotation are manually annotated summaries of the semantics of the feature. Max-Activating examples demonstrate the natural-language contexts that most activate these features, from Neuronpedia \citep{neuronpedia}.}
\label{tab:feature_table}
\end{table}

%% ARXIV: replace
% \begin{table}[t]
% \centering
% \input{Tables/extended_feature_descriptions}
% \caption{Descriptions of Matryoshka SAE features that often predict voxel responses. Feature \# corresponds to the dimension in the Matryoshka latent space. Feature Label and Annotation are manually annotated summaries of the semantics of the feature. Max-Activating examples demonstrate the natural-language contexts that most activate these features, from Neuronpedia \citep{neuronpedia}.}
% \label{tab:feature_table}
% \end{table}

From Fig.~\ref{fig:voxcat_pred}B, we see that \textbf{Abstract} and \textbf{Concrete} voxels are consistently supported by different feature subspaces. Moreover, many of these features are selected \textit{across} all eight participants (e.g., among the 160 voxels analyzed across eight participants---20 per subtype---feature 79 is selected in 156/160 abstract voxels; Fig.~\ref{fig:voxcat_pred}B). Thus, % out of a 32k-feature SAE basis, 
a small set of signed features is \textit{consistently} selected across individual brains. See Appendix~\ref{app:stability} for a robustness analysis.
From Table~\ref{tab:feature_table} it is evident that the interpretations of these features broadly align with existing interpretations \citep{tuckute2025two}. For example, \textbf{Abstract} voxels are driven by features related to emotions (Feat. \#40$+$; 144/160 voxels), and suppressed by features related to scenery (Feat. \#79$-$; 156 voxels). 

Furthermore, this analysis also serves to nuance existing interpretations of these voxel subtypes. By characterizing voxels based on their projection along a PC, prior work was limited to looking for high-level, symmetric interpretations that cleanly contrast voxel populations at both ends of a PC. By attempting to create voxel-level interpretations, our encoding models can reveal subtle asymmetries that were previously obscured. For example, we find an asymmetry when investigating a people/event-related feature (Feat. \#94), which suppresses \textbf{Abstract} voxels but is \textit{not} very reliable for predicting \textbf{Concrete} voxels. 
%Finally, the \textbf{Abstract} voxels are additionally driven by question- and consequence-related features (Feats. \#389$+$, \#1916$+$), as well as by failure/error-related content (\#16$+$).
Thus, what might appear to be a symmetric contrast when performing a PC-based analysis is, at the voxel level, asymmetric: some features drive one population without driving its counterpart to the same degree.

%%%%%%%%%%%%%%%%%%%%%%%%%%%%%%%%%%%%%%%%%%%%% STUDY 2 %%%%%%%%%%%%%%%%%%%%%%%%%%%%%%%%%%%%%%%%%%%%%

\section{Study 2: Identifying and Interpreting Uncharacterized Voxel Populations}
\label{Sec:Study2}

Having validated Augmented Sparse Encoding Models in Section~\ref{Sec:Study1}, we now present a case study of deriving novel hypotheses about the tuning of previously uncharacterized voxel populations. We use the voxels present in the \textbf{Ghost} subtype, as described in Section~\ref{methods:brain}. This population shows stable responses to language (i.e., high noise ceilings), but is not captured by the two main organizing PCs (Fig.~\ref{fig:brain_and_features}A). 
In this section, we use Augmented Sparse Encoding Models to tackle both the problem of \textit{identifying} coherent voxel subpopulations and then \textit{interpreting} them.

\begin{figure}
    \centering
    \includegraphics[width=\linewidth]{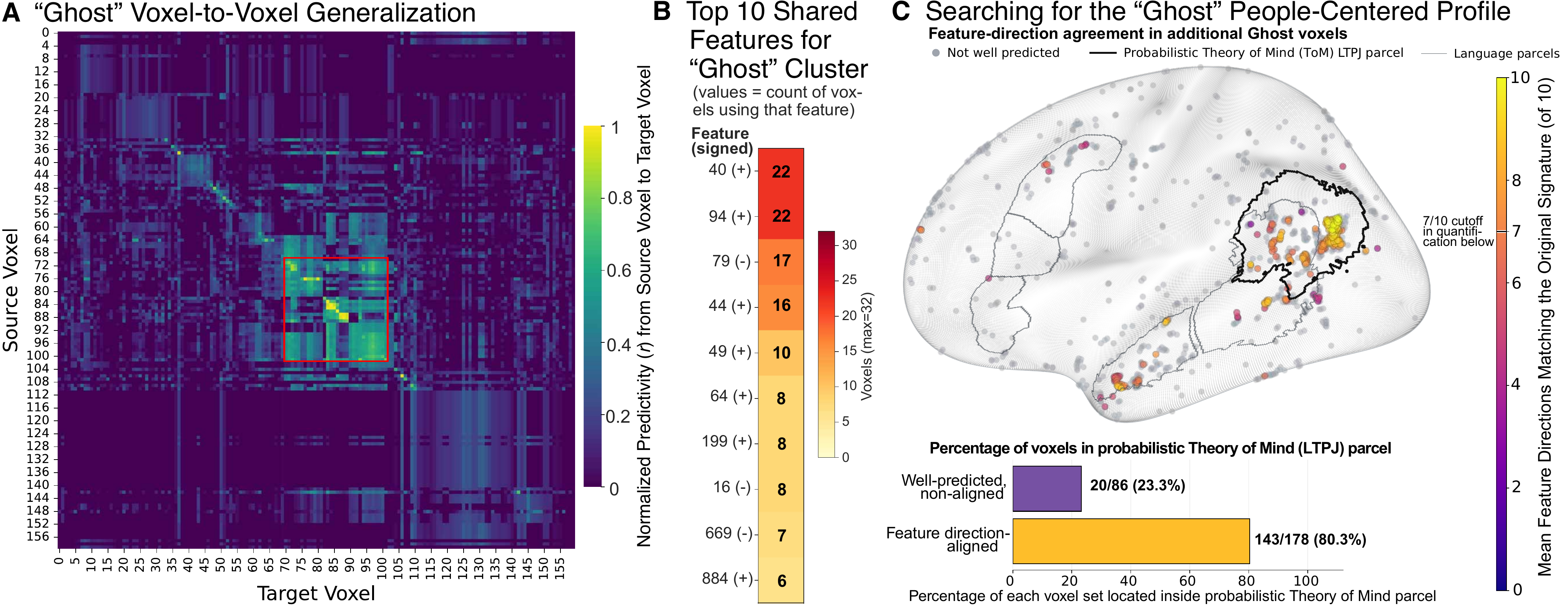}
    \caption{\textbf{(A)} Generalization heatmap for ``Ghost'' voxels (20 per participant, 160 in total), testing whether signed Matryoshka SAE features used to predict one voxel (source) generalize to predicting other voxels (target). Voxel order is determined by a hierarchical agglomerative clustering algorithm to cluster voxels with similar generalization profiles. Mutually well-predicted voxels are denoted by the red square. \textbf{(B)} Matryoshka SAE features selected by regressions on the red-square voxels in panel A. These features correspond to a ``people-centered'' voxel tuning. \textbf{(C)} We fixed the 10 signed features from panel B (i.e., the original Ghost cluster) and searched 1,490 additional Ghost voxels across participants. The surface map (fsaverage7; see individual maps in Appendix~\ref{app:ghost_search_ind}) overlays voxels from all eight participants, with well-predicted voxels colored by feature-direction with the 10-feature profile in panel B. The barplot quantifies the percentage of feature direction-aligned and well-predicted non-aligned voxels located within probabilistic theory of mind functional LTPJ \citep{dufour2013similar}.}
    \label{fig:ghost}
\end{figure}

\textbf{Analysis Methods}
To identify subpopulations of voxels that are predicted by a coherent set of features within the \textbf{Ghost} subtype ($G$), we assess how well features that are found to predict one voxel (a \textit{source} voxel, $s$) can generalize to predicting another voxel (a \textit{target} voxel, $t$). For every $s \in G$, we identify a signed sparse feature basis that predicts $s$ using all linguistic stimuli. Then, for every $t \in G$, we run a cross-validated and sign-constrained Ridge regression to predict $t$ using exactly the set of (signed) features that are used to predict $s$, ensuring that features are being used in ``the same way'' for both the source and target voxels (i.e., directional feature agreement). See Appendix~\ref{app:analysis_details} for further analysis details.

\textbf{Finding 1: A subset of the Ghost voxels form a coherent feature-defined subpopulation.}
Overall, many voxels in the \textbf{Ghost} subtype are not well predicted by surprisal or LM/SAE representations (Fig.~\ref{fig:ghost}A, and Appendix~\ref{app:ghost_reg}). However, one voxel subpopulation stands out as an exception: using the feature generalization analysis described above, we identify a coherent subpopulation of 32 voxels (Fig.~\ref{fig:ghost}A, red square). Within this subpopulation, signed features that predict one voxel also generalize to other voxels, consistent with a shared underlying feature basis.

\textbf{Finding 2: The coherent Ghost subpopulation is tuned to people-centered content.}
To interpret this voxel subpopulation, we fit Augmented Sparse Encoding Models to each voxel and examined the signed Matryoshka SAE features selected by the encoding models (i.e., the same analysis procedure as in Section~\ref{Sec:Study1}).
We find that the features used to predict this voxel subpopulation are typically driven by sentences that include ``people-centered'' features, such as discussions of relationships/emotions (Feat. \#40$+$), descriptions of people taking concrete actions (Feat. \#94$+$), or pronouns (Feat. \#49$+$). 
% Conversely, these voxels are suppressed by sentences that are describing the environment or scenery. 
% 199 is emotional expressions 
Next, we asked whether this feature profile extends beyond the original cluster. We fixed this cluster's 10 signed features (Fig.~\ref{fig:ghost}B) and searched 1,490 additional voxels satisfying the \textbf{Ghost} criteria (NCSNR $>0.4$ and correlations with both sentence PCs within $(-0.14,0.14)$). Of these voxels, 275 were well-predicted by a ridge regression using the 10 features (NC-normalized cross-validated threshold of $r > 0.5$). We visualized these voxels on the brain surface, coloring each voxel by the number of feature directions (out of 10) that matched the original signature (Fig.~\ref{fig:ghost}C). Strikingly, highly aligned voxels were strongly concentrated in the left temporoparietal junction (LTPJ)---a region implicated in Theory of Mind reasoning (\citealp{saxe2003people,miao2026common}). For quantification, we used the probabilistic Theory of Mind LTPJ parcel from \citet{dufour2013similar} and asked what proportion of feature-aligned voxels (defined as matching the original signature for at least 7 of the 10 feature directions, on average across cross-validation folds) fell within this parcel. In the five out of eight participants with more than three feature-aligned voxels, 143/178 voxels (80.3\%) fell within the LTPJ parcel, compared with 20/86 voxels (23.3\%) of well-predicted but non-aligned voxels (Fig.~\ref{fig:ghost}C). We emphasize that the signature generalizes beyond the original cluster to multiple participants despite no anatomical constraint on voxel selection, demonstrating how Augmented Sparse Encoding Models can recover coherent and interpretable representational structure shared across a subset of individuals.

\section{Study 3: Characterizing the Feature Basis of the Fronto-Temporal Language Network}
\label{Sec:Study3}

\begin{figure}
    \centering
    \includegraphics[width=\linewidth]{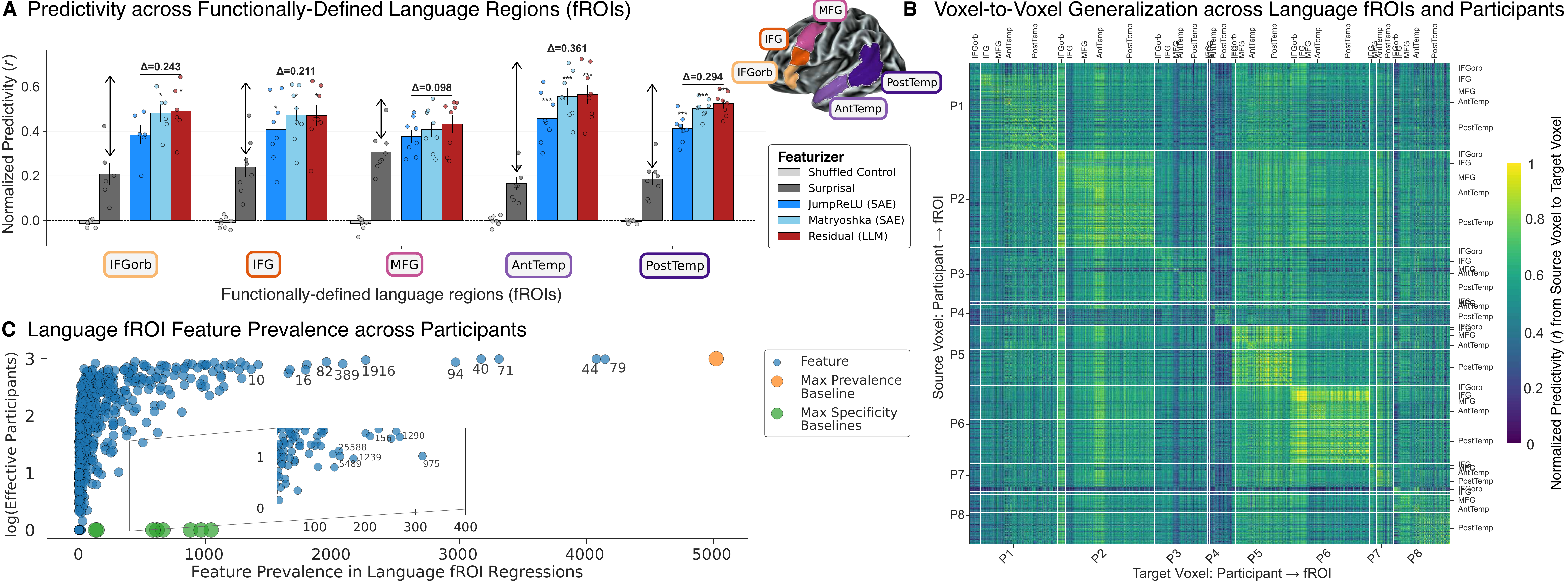}
    \caption{\textbf{(A)} Normalized encoding model predictivity of five language fROIs. \textbf{(B) }Generalization heatmap for language fROI voxels (2,296 voxels in total across eight participants)%; see Appendix~\ref{app:langfroigeneralization})
    , testing whether signed Matryoshka SAE features used to predict one voxel (source) generalize to predicting other voxels (target). Voxel order is sorted according to participant and fROI. \textbf{(C) }Quantification of the prevalence of features across participants. Each point is a feature (828 in total); the x-axis shows how often that feature is selected across voxel regressions, and the y-axis shows participant entropy in bits ($\log_2$ effective participants), so 
    higher values indicate broader sharing across individuals. Orange shows the theoretical maximum-prevalence baseline (a feature in every voxel regression), and green shows maximum-specificity baselines for features confined to a single participant (a feature in \textit{all} regressions from exactly \textit{one} participant).}
    \label{fig:langtuning}
\end{figure}
% (on average 287 voxels per participant, 2,296 in total)
% Language-network voxels share a common feature basis, but not uniformly. 

We next analyze the five canonical left-lateralized language regions \citep{fedorenko2024language}, using Augmented Sparse Encoding Models to ask whether the language network implements a largely shared representational code, or whether differences exist between regions and/or individuals.
% A central question is what language-selective areas represent and how they are organized: do these language-selective regions implement a largely shared representational code, or does more fine-grained substructure exist? 

% \red{Refer to lang-selectivity scatter stuff??}

\textbf{Finding 1: Language regions show mixed tuning to processing difficulty and content, but not uniformly across regions.}
First, in line with Section~\ref{Sec:Study1}, we find that Matryoshka SAEs still provide a better feature basis than JumpReLU SAEs, and are on par with residual-stream features (Fig.~\ref{fig:langtuning}A). 
Second, frontal fROIs are relatively well-captured by surprisal alone; in particular, the middle frontal gyrus (MFG) does not show a significant gain from SAE/residual-stream features beyond surprisal. In contrast, temporal fROIs show a significant boost from feature-based models above and beyond surprisal, indicating stronger sensitivity to content-linked structure during sentence comprehension (Fig.~\ref{fig:langtuning}A, $p<0.001$).
We replicated these findings in two independent 3T datasets with different participants, stimuli, and paradigms: sentence comprehension (\citealp{tuckute2024driving}; Appendix~\ref{app:3T}) and naturalistic story listening (Natural Stories; \citealp{de2025multilingual}; Appendix~\ref{app:rep_natstories}). In all three datasets, MFG shows the smallest gain over surprisal and AntTemp the largest.

\textbf{Finding 2: Language-network voxels draw on a common feature basis, with graded participant-specific variation.} 
Next, we ask whether the human language network is organized around a shared feature basis or more sharply separated regional and participant-specific feature spaces. 
First, we find that the voxel-to-voxel generalization heatmap shows substantial transfer between feature bases across the language network (Fig.~\ref{fig:langtuning}B), in stark contrast to the \textbf{Ghost} voxel-to-voxel heatmap (Fig.~\ref{fig:ghost}A). More specifically, we find that the heatmap does not reveal substantially higher transfer within-fROI than cross-fROI (within-fROI predictivity = 0.51 vs. cross-fROI predictivity = 0.50). This finding indicates substantial feature sharing across language regions, rather than region-specific feature bases. See Appendix~\ref{app:langfroigeneralization} for further analysis details.

Second, we also find substantial transfer across \textit{individuals}. Specifically, predictivity within-individuals is slightly higher than across individuals (within-participant predictivity = 0.61 vs. cross-participant predictivity = 0.49).
To understand the feature sharing across participants further, we quantify how broadly individual SAE features are shared across individuals by plotting each feature's prevalence across language-voxel regressions against the entropy of its distribution across participants (Fig.~\ref{fig:langtuning}C). 
Intuitively, high entropy means the feature occurs with roughly uniform frequency across participants; low entropy means it is concentrated in a few.
%Entropy is measured in bits, so $2^H$ gives the effective number of participants associated with that feature.
% This prevalence analysis revealed four regimes: a small number of features are extremely prevalent and broadly shared across all eight participants (upper-right portion of the plot; entropy $\approx$ 3). Their interpretations are given in Table~\ref{tab:feature_table} (see Appendix~\ref{app:froiqualitative} for further details). 
% Interestingly, many substantially less prevalent features are also reused across all eight participants despite appearing in far fewer voxel regressions. This suggests that the same type of heterogeneity \textit{within} individual brains occurs \textit{across} participants (i.e., these features are only predictive of a subset of voxels, but are found at equal rates across brains).
% A broad middle regime contains features with nontrivial prevalence reused across multiple---but not all---participants. Finally, a smaller low-entropy tail suggests participant-specific features (inset in Fig.~\ref{fig:langtuning}C).
This prevalence analysis revealed four regimes: a small number of features are extremely prevalent and broadly shared across all eight participants (upper-right portion of the plot; See Table~\ref{tab:feature_table} and Appendix~\ref{app:froiqualitative} for interpretations). %; entropy $\approx$ 3). 
% Their interpretations are given in Table~\ref{tab:feature_table} (see Appendix~\ref{app:froiqualitative}). 
Many substantially less prevalent features are also reused across all participants despite appearing in far fewer voxel regressions. This suggests that the same type of heterogeneity \textit{within} individual brains occurs \textit{across} participants, i.e., these features are only predictive of a subset of voxels, but are found at equal rates across brains.
A broad middle regime contains features with nontrivial prevalence reused across multiple participants, while a smaller low-entropy tail suggests participant-specific features (inset in Fig.~\ref{fig:langtuning}C).

% so then, what are these features? 
% most prevalent is 79 and 44 (above and beyond surprisal).
% 79 is about content; locations and scenes 
% 44 is about tokens/frequency: could signal lexical frequency? cause this is above sentence-level surprisal. so in this huge feature set, acknowledge the fact that e.g. this feature is clearly not about content, i.e. it is  language form (and also in spite of the sentences being auditorily presented, it’s not like they could see how many tokens a sentence was (same for feat 71). so despite including surprisal, there are still additional form-based features, i.e. there are traces of processing difficulty, possibly realted to lexical frequency (cite smth) and perhaps integration/WM demands (gibson) 
% 40 and 94 are also content, people/emotions.. 
% 1916 is Consequences & Statements, which could be related to processing difficulty 
% 389 is questions 
% 82 is technical terms (low freq words?)

What are these shared features? The two most prevalent features are Feat.~\#79 and \#44---Feat.~\#79 is content-linked, responding to locations and scenes, whereas Feat.~\#44 (and Feat.~\#71, another prevalent feature) appears to be driven by contexts with fewer tokens. Because each sentence contains the same number of words, this feature implicitly captures whether a sentence contains infrequent words that are divided into a larger number of tokens. 
%Because sentence-level surprisal was included explicitly in all models, 
These features may reflect additional form- or difficulty-linked features beyond average sentence-level surprisal alone, as surprisal is explicitly included in all models.
Other prevalent features captured people/emotions (Feat.~\#40), people/actions (Feat.~\#94), questions (Feat.~\#389), consequences/statements (Feat.~\#1916), and technical or specialized vocabulary (Feat.~\#82) (see Table~\ref{tab:feature_table}). These widely-shared features represent only a portion of a much richer feature basis: in total, 828 unique (unsigned) features were selected at least once across language-network voxels.
In summary, these results argue that the entire language network is organized around a shared, dominant feature set consisting of both content- and form-linked features. At a single-voxel level, this code consists of a small cross-participant core alongside a broader regime of features reused across only \textit{subsets} of participants.

%Taken together, these results argue that the entire language network is organized around a shared, dominant feature set consisting of both content- and form-linked features. Our results are consistent with recent views of the language network as an integrated system \citep{braga2020situating,fedorenko2024language}. We extend this view with our feature prevalence analysis, which shows that, at a single-voxel level, this code consists of a small cross-participant core alongside a broader regime of features reused across only \textit{subsets} of participants, likely reflecting individual variation in linguistic representations. 

\begin{figure}
    \centering
    \includegraphics[width=\linewidth]{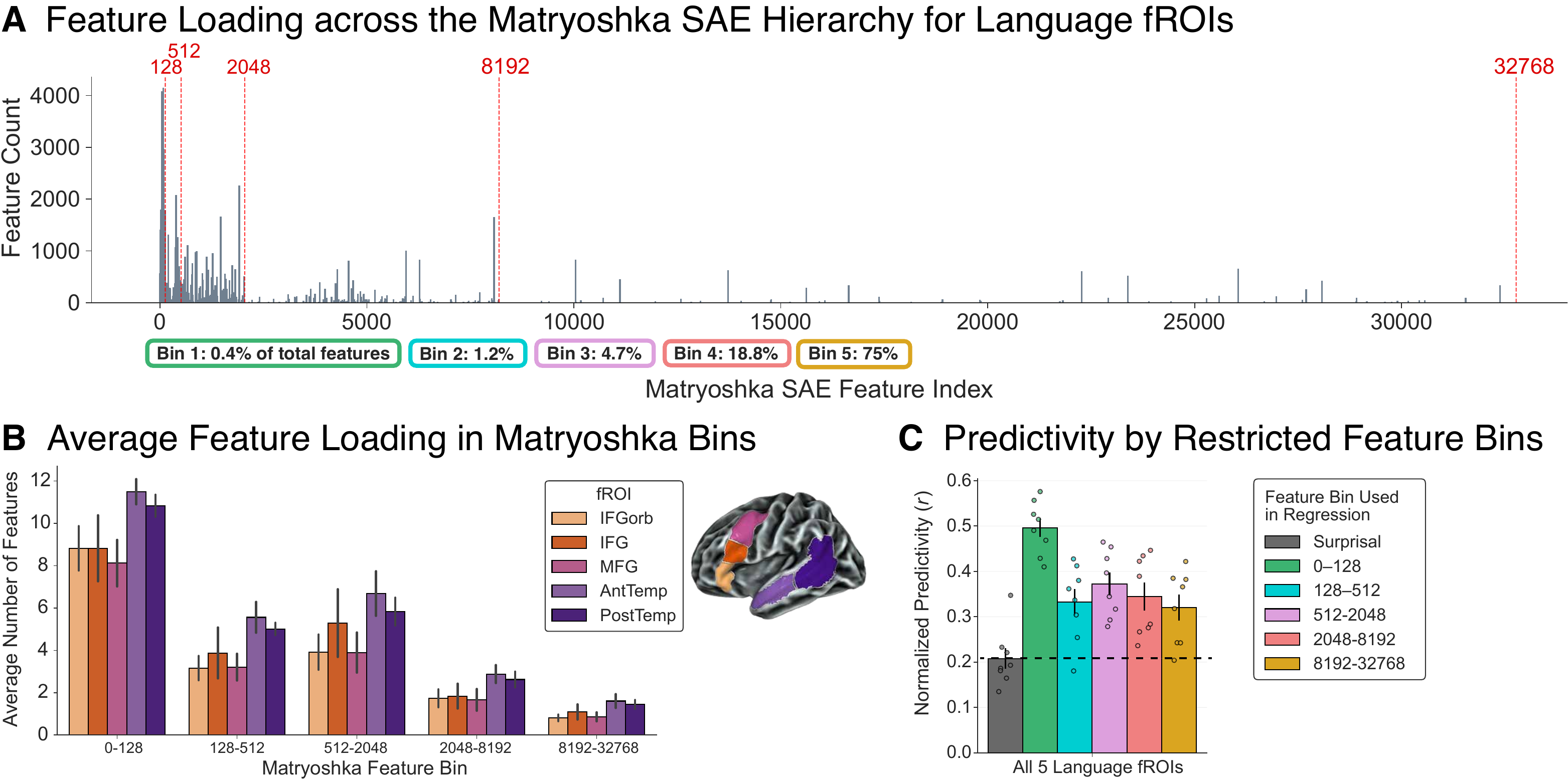}
    \caption{Augmented Sparse Encoding Models preferentially rely on general features to predict voxels across language fROIs, despite making up only 0.4\% of all features. \textbf{(A)} Histogram of how often each Matryoshka features (ordered by feature index) are selected by the encoding models. \textbf{(B) }Average number of features per Matryoshka bin. %when predicting voxels in each language fROI. 
    \textbf{(C) }Performance of encoding models with feature sets restricted to individual Matryoshka feature bins.}%General features are most predictive despite making up 0.4\% of all features.}
    \label{fig:granularity}
\end{figure}

\textbf{Finding 3: Human language network voxels are preferentially explained by general, widely-applicable LM features.}
Finally, we ask where brain-predictive features fall within the Matryoshka SAE hierarchy.
As described in Section~\ref{sec:Encoding}, the Matryoshka SAE imposes an ordered feature hierarchy: early latent features account for general, widely-applicable variance in LM representations \citep{bussmannlearning}. %and later features are progressively more granular 
We find that language-network encoding models preferentially recruit the most general Matryoshka features (feature indices $<128$; Fig.~\ref{fig:granularity}A,B,C, Appendix~\ref{app:rep_align}; see also Appendix~\ref{app:rep_natstories}). 
%A complementary analysis reaches the same conclusion: when models are fit using only a single Matryoshka feature bin, the most general bin outperforms every more granular bin, despite those bins containing far more features (Fig.~\ref{fig:granularity}C). 
%In Appendix~\ref{app:rep_align}, we further show that the most general bin also outperforms the union of all finer-grained bins. 
These findings indicate that brain-alignment with the human language network relies on the LM's general, widely-applicable features rather than with fine-grained, idiosyncratic features.
These findings are replicated using the PC-derived voxel subtypes (Appendix~\ref{app:rep_align}) and are not driven by a lack of active features in the finer-grained bins (Appendix~\ref{app:matryoshkastats}). 
These findings extend ``standard'' LM-encoding claims from predictivity to more specific correspondences between the properties of features that organize biological and artificial language systems.

\vspace{-1em}
\section{Discussion}
\label{sec:Discussion}
\vspace{-1em}
%\textbf{Conclusion}
Augmented Sparse Encoding Models move beyond prior approaches that treat both models and brains as black boxes.
Our empirical results show that small sets of features can be used to predict brain responses to language and separate processing difficulty-related signal from content. Within the language network, tuning is shared but not uniform: regions differ in their sensitivity to processing difficulty versus content, and content-linked features range from those broadly shared across individuals to individual-specific features.  
Below we discuss three implications of our findings: (i) the nature of linguistic representations in the language network, across regions and individuals (Section~\ref{app:discussion_language}); (ii) interpretation of the \textbf{Ghost} voxels that are tuned to people-related content (Section~\ref{app:discussion_ghost}); and (iii) which properties of LM representations mediate brain alignment with human language responses (Section~\ref{app:discussion_granularity}).

\subsection{The nature of linguistic representations: a shared feature basis}
\label{app:discussion_language}

%%%%%%%%%%%% LANG NETWORK
Prior work has shown that language-responsive regions show similar response profiles during controlled and naturalistic language paradigms \citep{rodd2010functional,fedorenko2020lack,shain2020fmri}. Here, we extend this view by demonstrating that these profiles rely on a common, interpretable feature basis. Small sets of \textit{signed} SAE features selected to predict one language-network voxel generalize strongly to voxels in other fROIs (Fig.~\ref{fig:langtuning}B), unlike the \textbf{Ghost} population (Fig.~\ref{fig:ghost}A). This finding indicates that features can be reused across language fROIs.
This full feature basis is large---predicting voxels across all five language fROIs draw on many features (828 in total)---but far from random: a handful of SAE features dominate, including ``processing difficulty-linked'' features that track token count (and,  implicitly, lexical frequency) and persist even with sentence-level surprisal as an explicit predictor (Appendix~\ref{app:froiqualitative}). 
Though a shared basis set of features can predict voxels across fROIs, different fROIs are \textit{not} uniformly tuned to these features. We find that processing difficulty features are useful for predicting voxels in all fROIs to different degrees (Fig.~\ref{fig:langtuning}A, surprisal bar), and that there exists a basis of content-linked features that are useful for predicting voxels in most, but not all, fROIs.

% there is a shared basis of processing difficulty which is broadly useful everywhere (finding 2 does not ask whether it “explains just a little or a lot of variance just like finding 1) (Fig.~\ref{fig:langtuning}A)
% and yes to “On top of that there is a shared basis of content features which are also broadly useful, but not helpful for every region”
% (Fig.~\ref{fig:langtuning}B)

Moreover, we find that a small set of core features also appear across \textit{individuals} (Fig.~\ref{fig:langtuning}C; in line with \citealp{tuckute2025two}). This core set of features is supplemented by many less prevalent features that are shared across most participants, as well as a broad set of features that are only shared within one or few individuals. While previous studies have shown that it is possible to generalize from one brain to another via LM residual-stream spaces (e.g., \citealp{de2025multilingual,tuckute2024driving}), these studies have been based on average fROI responses across the entire language network. The current analyses extend this prior work by generalizing at a single-voxel level instead of a network average (also see e.g., \citealp{tang2025semantic}), and further interpreting \textit{which} features are shared. These results bring us closer to understanding how linguistic meaning is represented in individual brains.
Future work with larger sets of stimuli can interpret these subset- and individual-specific features; one such direction is neural control \citep{bashivan2019neural,tuckute2024driving,antonello2024generative} in interpretable voxel subspaces, i.e., designing stimuli to drive targeted voxel populations.

% How is this feature basis shared across individuals? %% clunky
% While previous studies have shown that it is possible to generalize from one brain to another via LM residual-stream spaces (e.g., \citealp{de2025multilingual,tuckute2024driving}), these studies have been based on average fROI responses across the entire language network. The current work extends on this work by generalizing at a single-voxel level instead of a network average (also see e.g., \citealp{tang2025semantic}), and further interpret \textit{which} features are shared. Moreover, we characterize how \red{referring to the entropy stuff} (Fig.~\ref{fig:langtuning}C) these features are shared across participant, and we demonstrate a small set of dominant cross-participant features (in line with \citealp{tuckute2025two}) and a broader regime of features that are only shared within either a single individual or small subsets, bringing us closer to how linguistic meaning is represented in individual brains. 
% Future work with larger sets of stimuli can interpret these subset- and individual-specific features; one such direction is neural control \citep{bashivan2019neural,tuckute2024driving,antonello2024generative} in interpretable voxel subspaces, i.e., designing stimuli to drive targeted voxel populations.

\subsection{People-centered content outside the dominant language dimensions}
\label{app:discussion_ghost}

%%%%%%%%% GHOST / ToM
% \paragraph{People-centered content outside the dominant language dimensions.}
The Ghost analysis (Section~\ref{Sec:Study2}) shows that Augmented Sparse Encoding Models can be used to identify and interpret voxel subpopulations that exhibit coherent feature tuning. We first identified a coherent cluster of \textbf{Ghost} voxels predicted by a people-centered feature profile, including features related to relationships, pronouns, and actions involving people.

We next asked whether the same people-centered feature profile extended beyond this initially identified cluster. In an anatomically unconstrained search across 1,490 additional Ghost voxels, we recovered the profile in five of eight participants (see Appendix~\ref{app:ghost_search_ind}) and found that highly feature-aligned voxels were strongly concentrated in left temporoparietal junction (LTPJ). Within the five participants with more than three feature-aligned voxels, 80.3\% of feature-aligned voxels fell within a probabilistic functional LTPJ Theory-of-Mind parcel \citep{dufour2013similar} (\textbf{Fig.~\ref{fig:ghost}C}), compared with 23.3\% of well-predicted but non-aligned voxels. Importantly, no anatomical information was used to select these voxels. 

These results provide a suggestive link to areas typically implicated in social cognition and Theory-of-Mind reasoning \citep{rowe2001theory,stuss2001frontal,saxe2003people,dufour2013similar,miao2026common}. However, we emphasize that we did not include an independent Theory-of-Mind localizer in these participants, and therefore do not claim that these voxels are themselves functionally defined Theory-of-Mind regions. Similarly, failure to recover the profile clearly in the three out of eight participants could reflect either genuine inter-individual variability or limited sensitivity of the present dataset. We note that our analyses are restricted to the left hemisphere, where TPJ organization has been reported to be more variable across individuals than in the right hemisphere \citep{saxe2005making,dufour2013similar}.

This people-centered tuning profile also appears largely outside the core fronto-temporal language network. The strongest concentration of feature-aligned voxels lies adjacent to the probabilistic posterior temporal language parcel, rather than within it. This finding is consistent with prior work showing that the core language network is not tuned to social content or engaged by (non-verbal) theory-of-mind reasoning (\citealp{shain2023no,tuckute2024driving}, cf. \citealp{mellem2016sentence}). At the same time, a smaller proportion of voxels with the people-centric tuning profile fell within the two temporal language parcels (Fig.~\ref{fig:ghost}C), suggesting that social tuning during comprehension of short sentences appears partly adjacent to the core language network and partly interspersed with temporal language areas. Follow-up work using independent language and Theory-of-Mind localizers will be important for more clearly distinguishing these functional populations.

Finally, because this tuning is spatially distributed and not aligned with the dominant PCs, it would be tricky to detect from anatomy or low-dimensional response structure alone. Augmented Sparse Encoding Models isolate this people-centered voxel subpopulation through a small, signed feature basis, separating it from voxel populations tuned to processing difficulty or concrete--abstract content.

%reverse inference

% medial frontal area: in saxe&kanwisher + dufour
% so if this is a subpart of the TPJ network, we here demonstrate that these voxels do indeed show same feature tuning, aross these distinct areas, to short sentences, i.e. engagement of this ! 
% rowe 2001 stuss 2001 (lesions+neuroimaging) (frontal ROIs were less reliably localized in individuals)
% A nonverbal ToM contrasts does not engage core language regions \citep{shain2023no}. 

\subsection{General LM features predict brain responses}
\label{app:discussion_granularity}

%%%%%%%%%%%%%%% GRANULARITY
% \paragraph{General LM features predict brain responses.}
The Matryoshka hierarchy analysis (Section~\ref{Sec:Study3}) gives a more specific interpretation of why LM representations align with brain responses. The strongest alignment is \textit{not} carried by the multiplicity of fine-grained features that contribute to an LM's high-dimensional feature space. Instead, brain-predictive features are concentrated among the set of features that most generically characterize LM representations. This finding suggests partial overlap in the features that organize internal language representations in the human language network and LMs, even though these systems presumably have very different means of arriving at these representations. 

A priori, it was plausible that fine-grained features could have contributed substantially to predicting voxel responses. Various metrics that capture representation reconstruction steadily improve by incorporating such fine-grained features \citep{bussmannlearning}, and incorporating disentangled hierarchical features has resulted in improvements in sparse probing \citep{luo2026atoms}, and removing spurious correlations \citep{bussmannlearning}. However, our work suggests that most single voxels do not tend to represent fine-grained features, and are instead tuned to broader linguistic features. We do note that some fine-grained features predict responses in individual participants or small subsets of participants (Fig.~\ref{fig:langtuning}C, inset), likely reflecting individual-specific representations of linguistic meaning.

Our claim about general SAE dimensions aligns with \citealp{antonello2021low}, who showed that low-dimensional structure in LM representations is reflected in brain responses. We extend this work by identifying the relevant dimensions in an interpretable sparse feature basis. We further leverage these features to identify voxel subpopulations and characterize how these feature bases are shared across regions and individuals.

Finally, our SAE feature granularity analyses connect to work in the visual domain. For instance, \citet{longon2025superposition} show that SAE-based disentanglement can increase model--brain alignment when superposition obscures shared latent structure (also see \citet{mao2025sparse}). Here, we use a hierarchical SAE basis to ask \textit{which} levels of LM feature structure mediate brain alignment during language processing.

\paragraph{Limitations}
Augmented Sparse Encoding Models face several limitations stemming from the choice of SAE. Specifically, the interpretations that one can generate are limited to the feature basis of the SAE, and some SAE features remain opaque (especially for the JumpReLU SAE). However, other common critiques of SAEs (e.g., lack of causal relevance; \citealp{wuaxbench}) do not apply to our framework, as we merely require an unsupervised, interpretable feature basis to predict neural responses.
Other limitations stem from the neural data. Voxels encompass hundreds of thousands of neurons. It is possible that more general Matryoshka SAE features predict voxel tuning because voxels are a coarse-grained measure of neural activity, but one could imagine a coarse, yet ``brain-irrelevant'' feature set that fails to predict voxel responses, or could only do so with a larger and less semantically coherent feature set. Finally, collecting neural data for a larger set of diverse stimuli will enable further investigation of feature tuning during language processing; the present work provides a foundation for such analyses.

\section{Acknowledgments}
The authors would like to thank Thomas Serre, Ellie Pavlick, Andrea de Varda, Jacob Prince, Cindy Luo, Etha Hua, Eduardo Michelsen, and Jojo Yang for their helpful comments on an earlier version of the manuscript and for their many insightful discussions throughout the project. The authors thank Andrea de Varda for help with the naturalistic stories dataset. The authors would also like to thank Rachel Goepner for proofreading the manuscript. 

This material is based upon work supported by the National Science Foundation Graduate Research Fellowship under Grant No. 2439559. Any opinions, findings, and conclusions or recommendations expressed in this material are those of the author(s) and do not necessarily reflect the views of the National Science Foundation.
This work has been made possible in part by a gift from the Chan Zuckerberg Initiative Foundation to establish the Kempner Institute for the Study of Natural and Artificial Intelligence at Harvard University. We also acknowledge support from the McGovern Institute for Brain Research at MIT.
We also acknowledge funding from National Institutes of Health grant (NIH) R01EY034118.

\bibliography{iclr2027_conference}
\bibliographystyle{iclr2027_conference}

\pagebreak

%%%%%%%%%%%%%%%%%%%%%%%%%%%%%%%%%%%%%%%%%%%%%%%%%%%%%%%%%%%%

\appendix

\section{Detailed Voxel Selection Procedure}\label{app:voxel_selection}

We selected voxels of interest in two main ways: one based on the two principal components (PCs) of sentence-evoked responses identified by \citet{tuckute2025two}, and the other one based on whether a voxel is part of the fronto-temporal language network \citep{fedorenko2010new}.

\paragraph{PC-derived subtypes (Hard-to-Process, Easy-to-Process, Abstract, Concrete).}
We define these four subtypes via a three-step procedure which was applied separately for each of the eight participants. We selected $n = 20$ voxels per subtype.
 
\textit{Step 1: Identify voxels significantly predicted by either of the two PCs.}
We restrict to left-hemisphere voxels that are significantly predicted by a two-PC linear encoding model (PC\,1 and PC\,2 as regressors; cross-validated $R^2>1.06\%$; see \citealt{tuckute2025two} for details). This step ensures that the voxels entering subtype assignment are meaningfully captured by at least one of the two PCs.

\textit{Step 2: Assign voxels to subtypes based on PC correlations.}
For each remaining voxel, we compute its Pearson correlation with the sentence-level PC1 and PC2 scores (vectors of length 200, i.e., the number of stimuli in the experiment).
The threshold $|r| = 0.14$ marks the correlation values expected by chance for a two-tailed $p<0.05$ threshold based on 200 samples. We assign voxels to subtypes as follows:

\begin{table}[h!]
\centering
\renewcommand{\arraystretch}{1.3}
\begin{tabular}{l c c c}
\toprule
\textbf{Subtype} & \textbf{PC\,1 criterion} & \textbf{PC\,2 criterion} & \textbf{Ranking} \\
\midrule
Hard-to-Process  & $r_{\mathrm{PC1}} > 0$      & $|r_{\mathrm{PC2}}| < 0.14$ & top $n$ by $r_{\mathrm{PC1}}$ (descending) \\
Easy-to-Process  & $r_{\mathrm{PC1}} < 0$      & $|r_{\mathrm{PC2}}| < 0.14$ & top $n$ by $r_{\mathrm{PC1}}$ (ascending) \\
Abstract         & $|r_{\mathrm{PC1}}| < 0.14$ & $r_{\mathrm{PC2}} > 0$      & top $n$ by $r_{\mathrm{PC2}}$ (descending) \\
Concrete         & $|r_{\mathrm{PC1}}| < 0.14$ & $r_{\mathrm{PC2}} < 0$      & top $n$ by $r_{\mathrm{PC2}}$ (ascending) \\
\bottomrule
\end{tabular}
\caption{Criteria used to define the four voxel subtypes.}
\label{tab:subtypeselection}
\end{table}

\textit{Step 3: Ensure voxel reliability.}
We require an NCSNR of at least 0.4. If fewer than $n = 20$ voxels meet this threshold for a given subtype, we incrementally relax it in steps of 0.02 down to a minimum of 0.20 (thresholds: 0.40, 0.38, 0.36, \ldots, 0.20). This relaxation is occasionally necessary given that some participants have more reliable voxels than others (see the NCSNR values for each voxel subtype in Table \ref{tab:subtypestats}). No anatomical constraint is imposed beyond restricting to the left hemisphere. Across all subtypes, the $n \times 4 = 80$ selected voxels are mutually exclusive (i.e., no voxel appears in more than one subtype).
 
\paragraph{Ghost voxels.}
Ghost voxels are selected separately and do \textit{not} require significant prediction by the two-PC encoding model, since by definition they are not well captured by either PC. Starting from all left-hemisphere voxels with NCSNR $> 0.4$, we identify voxels whose correlations with both PC\,1 and PC\,2 fall within $(-0.14, \; 0.14)$ (i.e., chance-level). Among these voxels, we rank by the sum of absolute correlations ($|r_{\mathrm{PC1}}| + |r_{\mathrm{PC2}}|$) and select the $n = 20$ voxels with the lowest combined absolute correlation, i.e., those closest to the origin in PC-correlation space (see Figure~\ref{fig:brain_and_features}A).
 
\paragraph{Language-network fROIs.}
Using a standard approach in cognitive neuroscience, we use an independent ``functional localizer'' experiment to select voxels that are responsive to language \citep{saxe2006divide,fedorenko2010new}. For each participant, we identify voxels falling within five predefined left-hemisphere parcels---IFG, IFGorb, MFG, AntTemp, and PostTemp---and select the top $10\%$ of voxels by $t$-statistic from a standard \textit{sentences}~$>$~\textit{nonwords} localizer contrast \citep{fedorenko2010new}. As in all other analyses, we require NCSNR $> 0.4$ for the language fROI voxels.

\begin{table}[ht]
\centering                                                              
\caption{Summary statistics for the five voxel subtypes across eight participants.}        
\begin{tabular}{lccc}                                       
\toprule        
Subtype & NCSNR & Lang $t$-stat & $N$ voxels/participant \\
\midrule        
Hard to process & $0.558 \pm 0.012$ & $1.874 \pm 0.660$ & $20.0 \pm 0.0$ \\
Easy to process & $0.366 \pm 0.026$ & $-0.467 \pm 0.177$ & $20.0 \pm 0.0$ \\
Abstract & $0.515 \pm 0.016$ & $5.292 \pm 0.697$ & $20.0 \pm 0.0$ \\
Concrete & $0.490 \pm 0.036$ & $2.522 \pm 0.343$ & $20.0 \pm 0.0$ \\    
Ghost & $0.428 \pm 0.005$ & $1.045 \pm 0.312$ & $20.0 \pm 0.0$ \\  
\bottomrule
\end{tabular}
\label{tab:subtypestats}
\end{table}                                                                                       
\begin{table}[ht]
\centering
\caption{Summary statistics for language fROI voxels (NCSNR $> 0.4$) across eight participants.}
\label{tab:langfroi_summary}
\begin{tabular}{lccc}
\toprule
fROI & NCSNR & Lang $t$-stat & $N$ voxels/participant \\
\midrule
IFGorb & $0.474 \pm 0.008$ & $6.402 \pm 0.823$ & $74.7 \pm 13.3$ \\
IFG & $0.472 \pm 0.009$ & $5.935 \pm 0.931$ & $62.6 \pm 17.7$ \\
MFG & $0.463 \pm 0.015$ & $6.844 \pm 0.740$ & $90.8 \pm 26.2$ \\
AntTemp & $0.478 \pm 0.009$ & $7.518 \pm 0.743$ & $74.5 \pm 18.1$ \\
PostTemp & $0.486 \pm 0.011$ & $9.000 \pm 0.738$ & $343.8 \pm 66.1$ \\
\bottomrule                                              
\end{tabular}
\label{tab:langfroistats}
\end{table}                                                                                                                     
\section{Replication Using \texttt{gemma-2-2b} Layer 14}
\label{app:layer14replication}

\begin{figure}
    \centering
    \includegraphics[width=\linewidth]{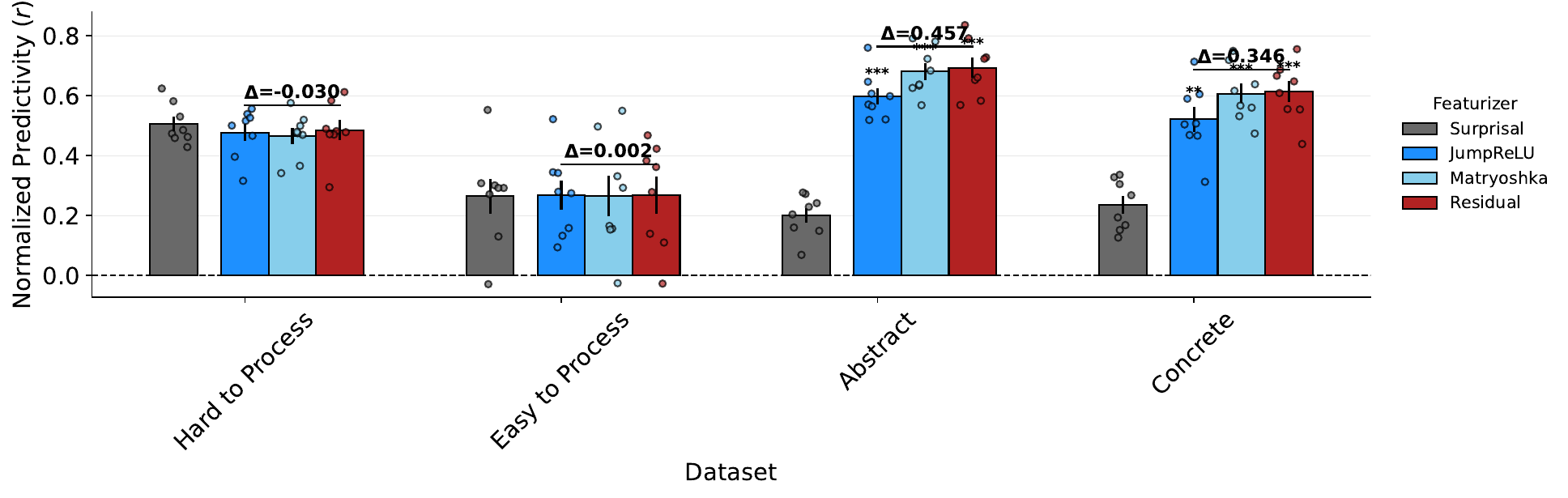}
    \caption{Replication of Fig.~\ref{fig:voxcat_pred}A using representations from layer 14 of \texttt{Gemma-2-2b}}
    \label{fig:Layer14}
\end{figure}

In this section, we replicate our primary quantitative regression results in Section~\ref{Sec:Study1} using LM representations from layer 14 (rather than layer 12) of \texttt{gemma-2-2b}. From Fig.~\ref{fig:Layer14}, one can see that our results are qualitatively similar to those in Fig.~\ref{fig:voxcat_pred}A.

\section{Additional Methodological Details for All Analyses}
\label{app:analysis_details}

\paragraph{Computational Costs} Every analysis presented in this work was performed using CPUs. A single Nvidia TitanRTX GPU was used to featurize stimuli.

\paragraph{Cross-Validated Predictivity Analysis}

For each voxel, we use 5-fold cross validation over sentence stimuli to estimate the performance of our encoding models in Sections~\ref{Sec:Study1} and ~\ref{Sec:Study3}.
For each train/test split, we employ a two stage processing pipeline. The first stage consists of feature selection. For Matryoshka and JumpReLU feature sets, we first filter to the  8000 most promising features using F-tests from univariate linear regression on each feature. This removes features that do not correlate with voxel responses. Next, we fit an L1-penalized (LASSO) regression. We use a nested 5-fold CV to identify the LASSO $\alpha$ hyperparameter, searching over a log-spaced range of 10 values from .01 to 1. After identifying the $\alpha$ hyperparameter, we fit the LASSO regression to the entire train split to identify a sparse feature basis.

The second stage consists of fitting a generalizable regression model to the sparse feature basis identified in the first stage. We employ a Ridge regression for this purpose. Regardless of the results of the LASSO feature selection, we always include the surprisal feature in the Ridge regression. This procedure ensures that the feature set is never empty, and allows for clean comparisons between full LM encoding model predictivity and a surprisal-only baseline. We use a nested leave-one-out cross validation to identify the optimal $\alpha$ hyperparameter, searching over a range from .01 to 100000. We refit the Ridge regression on the entire train split using the optimal alpha in order to predict test split responses.

Within each fold, brain responses were $z$-scored using training-set statistics only, and surprisal was $z$-scored using the training set only, preventing information leakage across folds.
The final predictivity score was defined as the mean Fisher-$z$-transformed correlation across folds, noise-ceiling normalized at the voxel level. 

For testing statistical significance among different featurizers (Figs.~\ref{fig:voxcat_pred}, \ref{fig:langtuning}, \ref{fig:Layer14}, \ref{fig:GhostRegression}), we use a paired $t$-test on the predictivity values obtained across all eight participants. For each figure, * = $p<.05$, ** = $p<.01$, *** = $p<.001$

\paragraph{Generalization Analysis}

Generalization analyses described in Sections~\ref{Sec:Study2} and~\ref{Sec:Study3} proceeded as follows. At a high level, the goal is to understand which voxels in a population are predicted by the same sets of features. To do this, we first identify a sparse set of features that predict one voxel, a \textit{source voxel}, and then assess how well the identified (signed) feature set predicts each other voxel in the population. 

In service of this goal, we identify features that predict a source voxel by first filtering features using an F-Test, then fitting a LASSO regression for feature selection, and finally fitting a Ridge regression to arrive at a final set of coefficients and features. Regardless of the results of the LASSO feature selection, we always include the surprisal feature in the Ridge regression. We run this pipeline using all sentence stimuli, resulting in a sparse, signed set of features that predict the source voxel.

Next, we wish to assess the degree to which this sparse, signed feature set can predict each of the other voxels in the population. We call each of these voxels \textit{target} voxels. We use 5-fold cross validation over sentence stimuli to assess how well the (signed) features predict a target voxel. Within each of these folds, we use an 80-20 train/test split (rather than a computationally expensive nested 5-fold CV) to identify the optimal $\alpha$ value for a sign-constrained Ridge regression, searching over a range from .01 to 100000. After identifying the optimal $\alpha$ value for a fold, we refit a sign-constrained Ridge regression and predict test split voxel responses.

\paragraph{Feature Interpretation Analysis}

The feature interpretation analyses employed by Sections~\ref{Sec:Study1}, \ref{Sec:Study2}, and \ref{Sec:Study3} proceed by first filtering features using an F-Test, then fitting a LASSO regression for feature selection, and finally fitting a Ridge regression to arrive at a final set of coefficients and features. To arrive at a feature set that predicts voxel responses to all stimuli, we run this pipeline over all sentence stimuli (rather than performing a cross-validated analysis). After fitting the Ridge regression, we note the sign of each selected feature.

\paragraph{Extended Ghost Voxel Analysis}
In Section~\ref{Sec:Study2}, we identified a set of signed Matryoshka features that seemed to correspond with ``people-centered'' content and predicted a subpopulation of \textbf{Ghost} voxels. After identifying this signature, we attempted to search for more Ghost voxels that were well-predicted by the same signed features. To do this, we fit cross-validated ridge regressions using these 10 features on an additional 1,490 \textbf{Ghost} voxels. In contrast with the other regressions in this work, we omitted the surprisal feature, and performed no feature filtering (i.e., F-test or LASSO regressions). We then performed a post-hoc analysis of the signs of the coefficients associated with each of these features, in order to see whether they aligned with the predefined signature of ``people-centered'' features.

\paragraph{Comparison to Existing Approaches: Additional Methods}
Here we provide technical details on our comparison to existing methods in Fig.~\ref{fig:brain_and_features}C. For \citet{zeng2025disentangling}, we featurized stimuli using their pretrained sparse concept featurizer dictionary, which contains 1000 sparse features. For \citet{benara2024crafting}, we used the 29 questions for their final reported fMRI model (from their larger 674-question candidate pool). Following their ensemble procedure, each (sentence, question) pair was evaluated using \texttt{Mistral-7B-Instruct-v0.2} zero-shot and \texttt{Llama-3-8B-Instruct} with both zero-shot and five-shot prompts. Each configuration produced a binary yes/no response, and the three responses were averaged to produce one 29-dimensional QA feature vector per sentence. Finally for Eng1000, we created one average feature vector per sentence using the 985 features from \citet{huth2016natural}.

After generating these different feature sets, we evaluate each using the same regression pipeline employed for the Augmented Sparse Encoding Models: F-test filtering, then LASSO filtering and ridge regression.

\newpage
\section{Feature Set Support Size}
\label{app:support_size}

\begin{figure}
    \centering
    \includegraphics[width=0.6\linewidth]{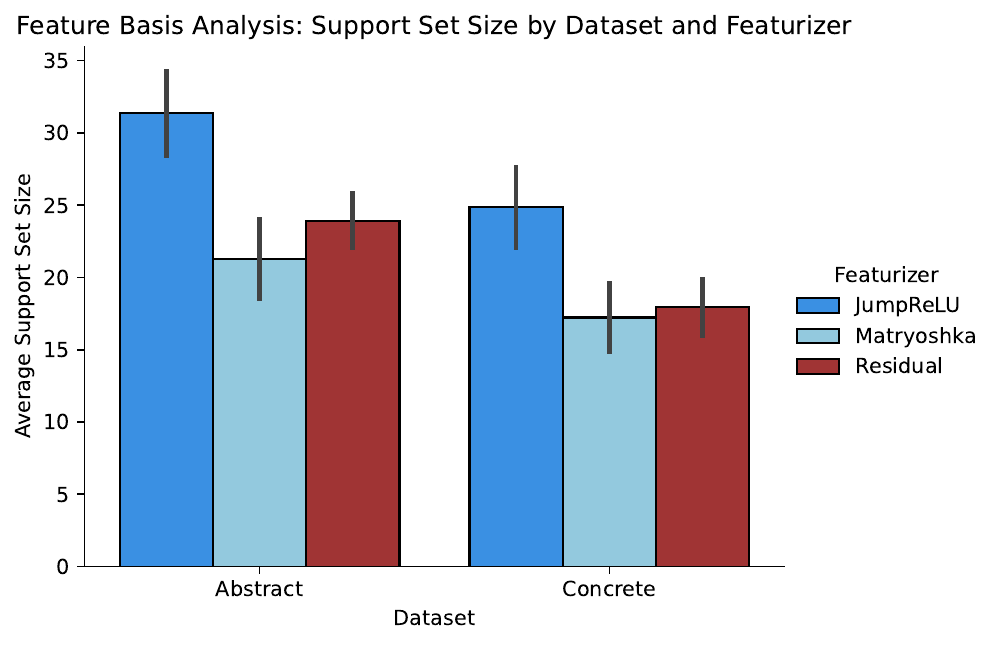}
    \caption{Average support set sizes (number of features selected by LASSO regression) for \textbf{Abstract} and \textbf{Concrete} voxel subtype regressions for all three feature sets used in Section~\ref{Sec:Study1}.}
    \label{fig:subtype_support}
\end{figure}

\begin{figure}
    \centering
    \includegraphics[width=0.6\linewidth]{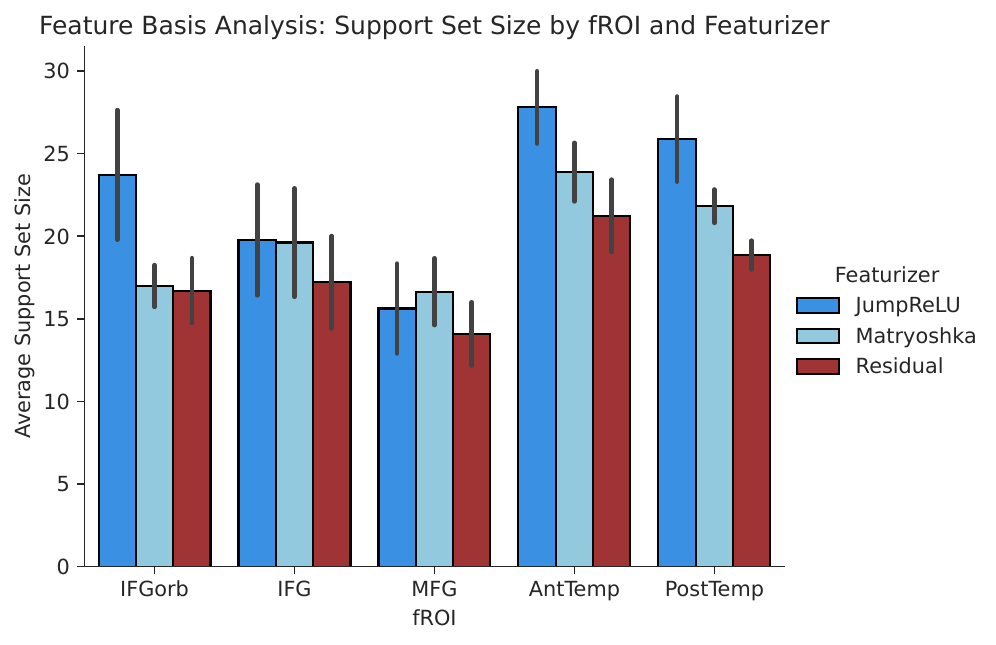}
    \caption{Average support set sizes (number of features selected by LASSO regression) for fROI regressions for all three feature sets used in Section~\ref{Sec:Study3}.}
    \label{fig:froi_support}
\end{figure}

In Fig.~\ref{fig:subtype_support}, we present the average support size (i.e., number of features) resulting from LASSO-based feature selection using residual stream, JumpReLU, and Matryoshka feature sets for \textbf{Abstract} and \textbf{Concrete} voxel subtypes. When including processing difficulty voxel subtypes, the average Matryoshka support size is 18 features. In Fig.~\ref{fig:froi_support}, we do the same, except for each language fROI. We find that Matryoshka feature sets result in comparable support set sizes to residual stream regressions, and typically smaller support sets than JumpReLU regressions.

\newpage
\section{Ghost Voxel Regressions}
\label{app:ghost_reg}
\begin{figure}
    \centering
    \includegraphics[width=.5\linewidth]{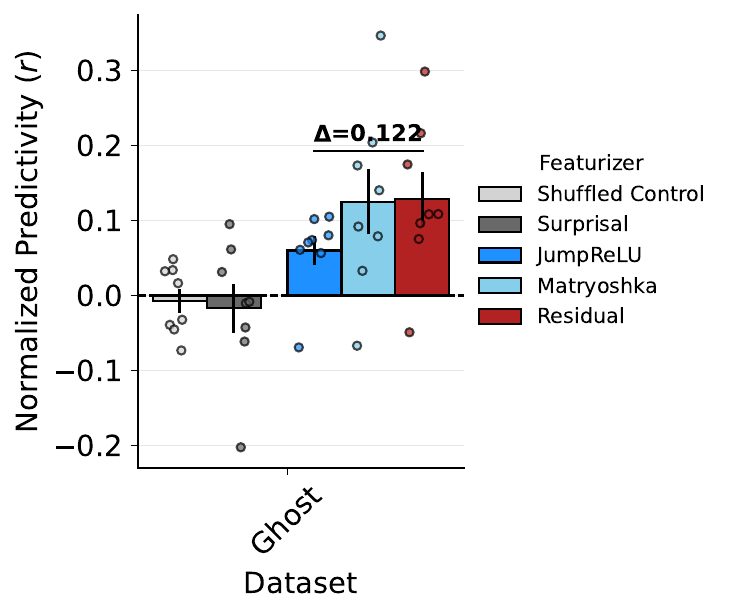}
    \caption{Regression predictivity for Ghost voxels using all feature sets.}
    \label{fig:GhostRegression}
\end{figure}

In this section, we demonstrate that, on average, none of our feature sets can predict voxels in the \textbf{Ghost} subtype well. From Fig.~\ref{fig:GhostRegression}, one can see that even the best performing feature sets (Residual and Matryoshka) only achieve Normalized Predictivity of approximately .1.

\section{Split-Half Qualitative Analysis}\label{app:stability}
In this section, we perform a split-half analysis to assess the stability of the qualitative findings presented in Figure~\ref{fig:voxcat_pred}. Specifically, we split our stimuli into two disjoint sets, and reran the qualitative analysis presented in Figure~\ref{fig:voxcat_pred} on each set independently. We identify the top 15 features in each split (identically to Figure~\ref{fig:voxcat_pred}), and find that their IoU is .58, with 19 features in the union. For each dataset in {abstract, concrete}, we then considered these 19 features and counted, separately within each split, the number of voxels for which each feature was selected. These feature selection frequencies were highly correlated across splits (r=.87 for abstract, r=.85 for concrete). Furthermore, the mean per-voxel IoU between splits is .24 for Abstract and .19 for Concrete. Despite there being more than 32K distinct features available to the model, both the aggregated feature-selection patterns and the individual voxel feature-selection patterns were reproducible across stimulus set splits.

\newpage
\section{Participant-level Surface Maps for Additional Ghost Voxel Analyses}
\label{app:ghost_search_ind}

\begin{figure}
    \centering
    \includegraphics[width=.99\linewidth]{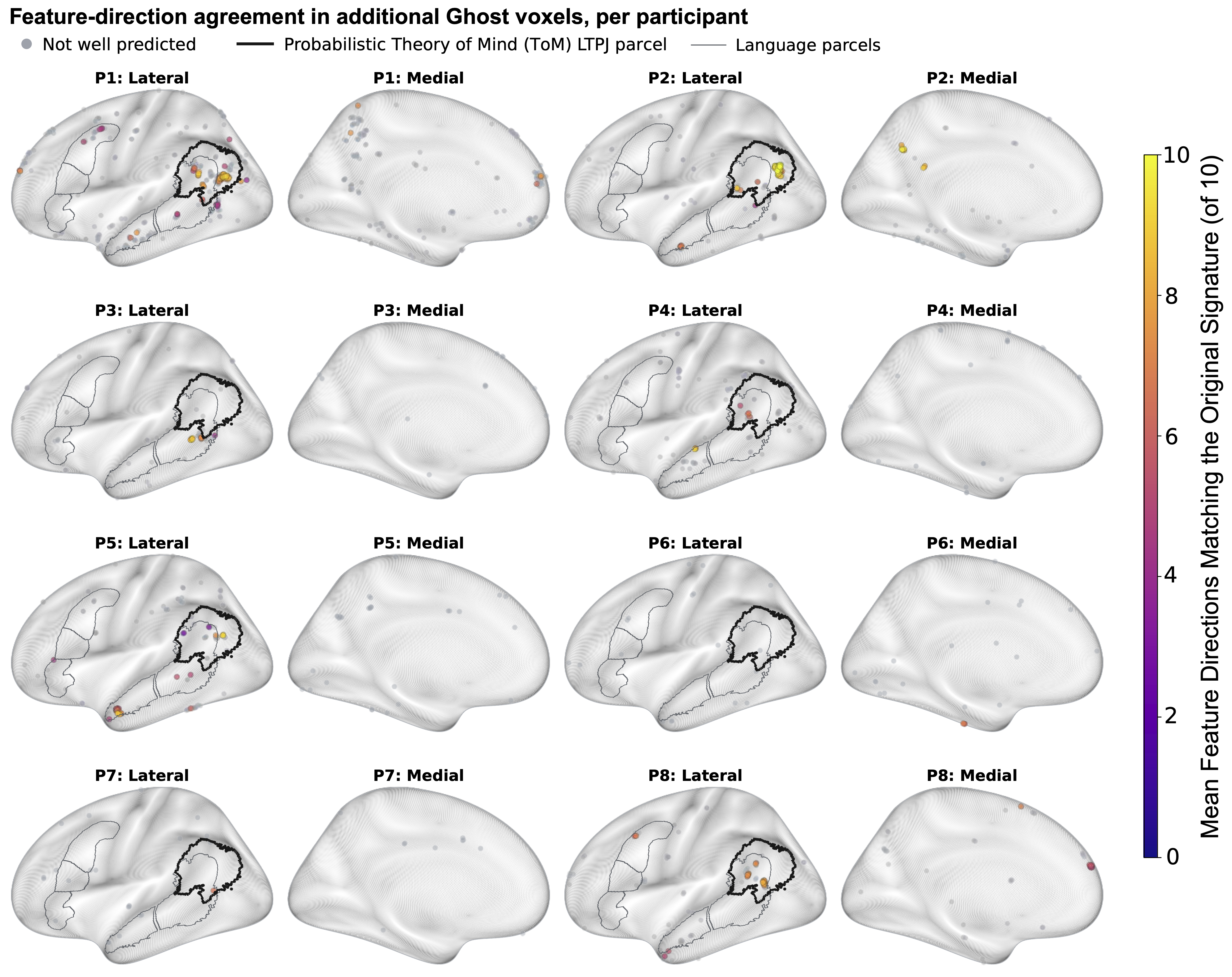}
    \caption{Participant-level maps from the additional \textbf{Ghost} voxel search (complementing the group map in \textbf{Fig.~\ref{fig:ghost}C}). Well-predicted voxels are colored by the number of feature directions (out of 10) matching the original people-centered signature; gray voxels were not well predicted. Black outlines indicate the probabilistic Theory-of-Mind LTPJ parcel \citep{dufour2013similar} and gray outlines indicate the language parcels.}
    \label{fig:ghost_search_ind}
\end{figure}

We provide participant-level maps from the \textbf{Ghost} voxel search in Figure~\ref{fig:ghost_search_ind}. For anatomical quantification and visualization, we projected the released binary LTPJ mask \citep{dufour2013similar} from MNI152 volume space to the fsaverage7 surface using \texttt{neuromaps} \citep{markello2022neuromaps}, with nearest-neighbor interpolation. 

\section{Replication Using 3T Sentence-Level Brain Dataset}\label{app:3T}

In this section, we replicate our main quantitative results from Section~\ref{Sec:Study3} using an independent sentence fMRI dataset \citep{tuckute2024driving}. Five participants read 1,000 linguistically diverse sentences during 3T fMRI. Language fROIs were defined using the same localizer procedure as in the main analyses, and we refer the reader to \citet{tuckute2024driving} for additional methodological details on this dataset.
This dataset differs from the main 7T dataset in several respects: different participants, lower field strength (3T vs. 7T), different stimulus modality (reading vs. listening), different MRI preprocessing, and a different experimental design. 

The experiment was designed for fROI-level rather than voxel-level analyses: it was acquired at 3T (vs. 7T) and contains no within-participant stimulus repetitions, so voxel-wise reliability cannot be estimated from repeated stimulus presentation \citep{allen2022massive}.

\begin{figure}[h]
    \centering
    \includegraphics[width=\linewidth]{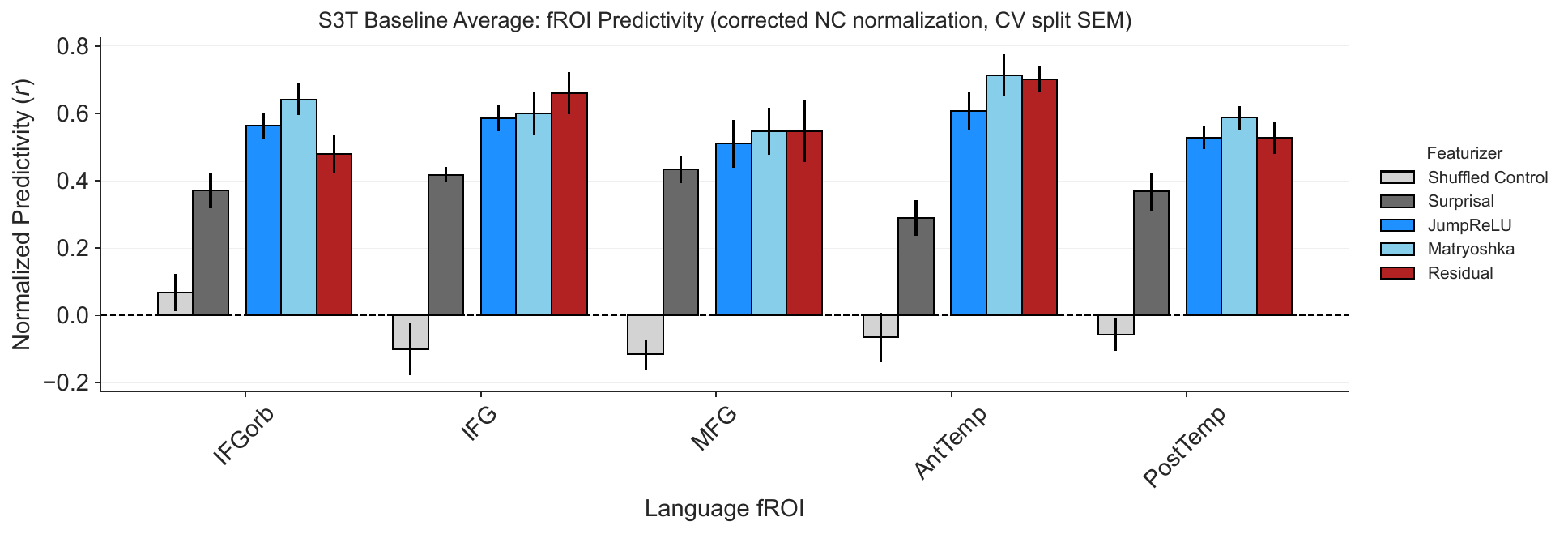}
    \caption{Replication of Fig.~\ref{fig:langtuning} on an independent dataset. Following the procedure in \citet{tuckute2024driving}, the brain responses in each language fROI were averaged across the five participants. Predictivity is reported as the mean Fisher-$z$-transformed correlation across 5 CV folds, normalized by the across-participant noise ceiling; see \citet{tuckute2024driving} for details. Error bars show SEM across 5 cross-validation splits.}
    \label{fig:3TReplication}
\end{figure}

Despite these differences, Fig.~\ref{fig:3TReplication} shows the same qualitative pattern as Fig.~\ref{fig:langtuning}: MFG exhibits the smallest difference between surprisal vs. other feature sets, and AntTemp exhibits the largest difference.

\section{Replication Using 3T Naturalistic Stories Brain Dataset}\label{app:rep_natstories}

In this section, we replicate our main quantitative results from Section~\ref{Sec:Study3} using the Natural Stories fMRI dataset used by \citet{de2025multilingual} (expanded from \citet{blank2014functional}).
The dataset comprises nine approximately five-minute spoken English stories \citep{futrell2018natural,paunov2022differential}. We used the publicly released group-averaged response time series for five functionally defined left-hemisphere language fROIs: IFGorb, IFG, MFG, anterior temporal, and posterior temporal regions. Predictors were aligned to the released 2-s response bins, and we used the released response series with a three-TR hemodynamic shift for modeling (see details in \citet{de2025multilingual}).

For feature extraction, we extracted \texttt{gemma-2-2b} layer-12 residual-stream representations and the corresponding Matryoshka and JumpReLU SAE features. Features were computed using a trailing 100-word causal context; subtoken representations were averaged within words and word representations within each 2-s bin. As in the main text analyses, we compared surprisal alone with models that augmented surprisal with selected residual-stream or SAE features.

Model performance was evaluated with nine-fold leave-one-story-out cross-validation. Within every outer fold, feature screening, feature selection, and hyperparameter selection used only the eight training stories. Candidate model features were first reduced to the 8,000 features with the strongest univariate associations and then selected using L1 regression with leave-one-training-story-out cross-validation. In line with our framework, surprisal was retained in each of the final models. The selected predictors were refit using ridge regression, with the ridge parameter chosen through an inner leave-one-story-out loop over the training stories. To minimize story-onset effects, the first 50 response bins (100 s) of each story were excluded during feature selection and inner and outer scoring to avoid ``ramping-bias'' (see \citet{antonello2023scaling}); the final ridge model itself was fit using all available bins from the training stories. Predictivity was quantified as the Pearson correlation between predicted and observed responses in the held-out story.

\begin{figure}[h]
    \centering
    \includegraphics[width=\linewidth]{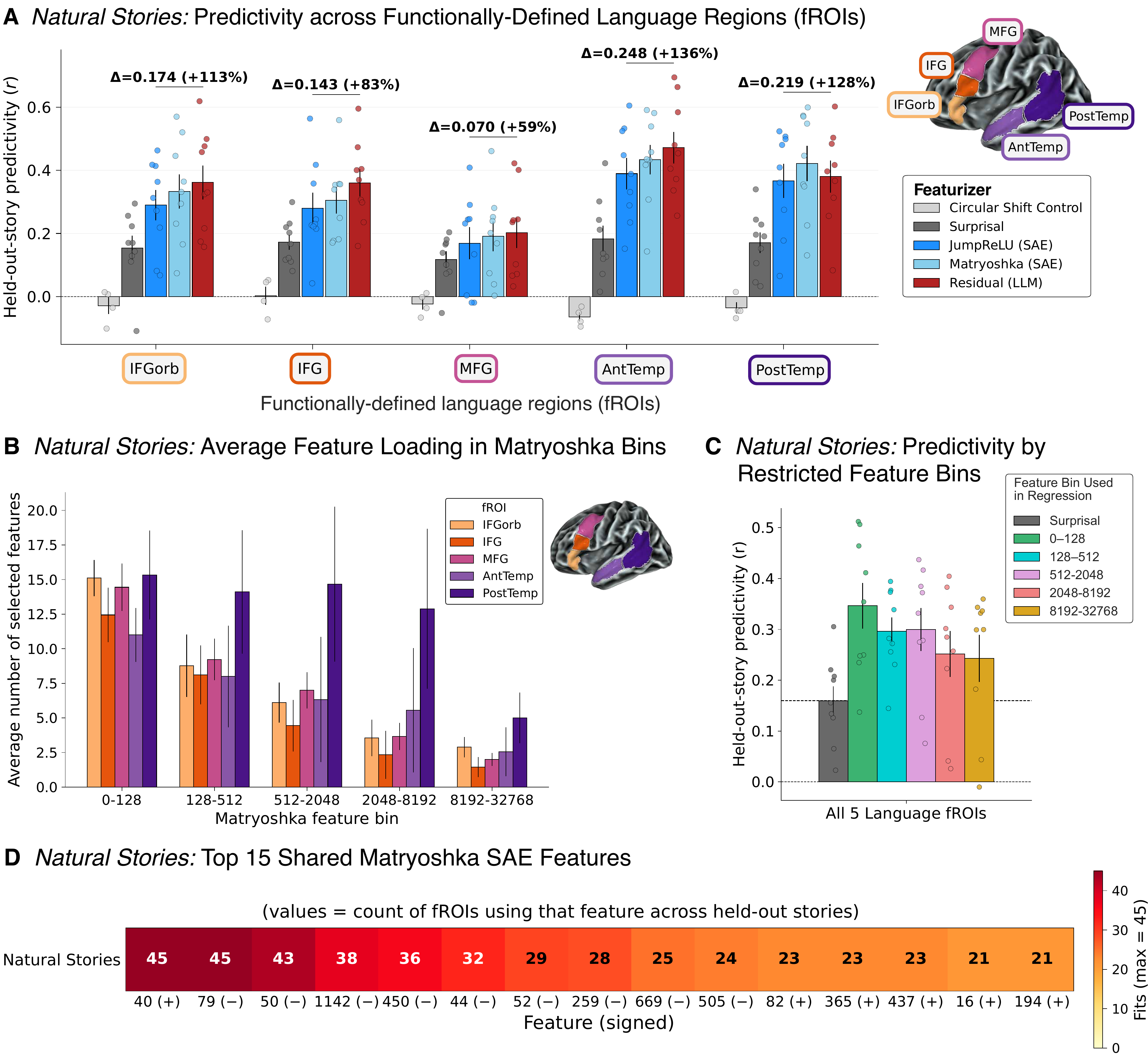}
    \caption{Replication of the language-network findings in the Natural Stories dataset, using naturalistic language comprehension. \textbf{(A)} Predictivity across language fROIs, replicating Fig.~\ref{fig:langtuning}A. As a control, we used a circular shift baseline \citep{de2025multilingual}. For the actual models, dots denote held-out stories, bars show mean Pearson correlation, and error bars show SEM across nine stories. For the control, dots denote the four shift-specific means across stories; bars and error bars show their mean and SEM. The $\Delta$ annotations report the raw and percentage increase from surprisal to the average predictivity of the three model-feature representations.
    \textbf{(B--C)} Matryoshka feature loading and restricted-bin predictivity, replicating Fig.~\ref{fig:granularity}B--C. Individual dots in Panel C denote individual held-out-story folds, bars denote their mean, and error bars show the standard error of the mean. \textbf{(D)} Most frequently shared Matryoshka features across language fROIs, complementing the feature-prevalence analysis in Fig.~\ref{fig:langtuning}C. Cell color indicates the number of fROI-by-story fits in which each signed feature was selected. Five fROIs were used: IFGorb, IFG, MFG, anterior temporal, and posterior temporal regions.}
\label{fig:natural_stories_replication}
\end{figure}

Despite major differences in experimental stimuli and methodological details, Fig.~\ref{fig:natural_stories_replication} replicates the key quantitative patterns from our main analyses: sparse SAE features substantially improve prediction over surprisal alone and the temporal fROIs show a larger boost from ``content-based'' feature predictors compared to frontal ones (e.g., MFG)  (\textbf{Fig.~\ref{fig:natural_stories_replication}A}). The predictivity of the SAEs (especially Matryoshka) approaches the predictivity level of residual stream features (\textbf{Fig.~\ref{fig:natural_stories_replication}A}). Moreover, coarse, earlier Matryoshka feature bins are preferentially selected in the regressions and more predictive (\textbf{Fig.~\ref{fig:natural_stories_replication}B--C}). Finally, a small set of features is consistently selected across language fROIs (\textbf{Fig.~\ref{fig:natural_stories_replication}D}), and the top-loading features are shared with those in the main 7T dataset (see \textbf{Fig.~\ref{fig:froiqualitative}} and \textbf{Fig.~\ref{fig:langtuning}}).

\section{Detailed Information about Language fROI Voxel Selection for Generalization Analysis}\label{app:langfroigeneralization}

Due to the computational cost of running the voxel-to-voxel generalization analyses for the language fROI voxels (which grows quadratically with the number of voxels; Section~\ref{Sec:Study3}), we sub-sampled fROI voxels (see Table~\ref{tab:langfroistats}). The total number of language fROI voxels across participants was 5{,}021 voxels. The sub-sampling selection proceeds as follows:
First, we ``deduplicated'' voxels with identical brain response profiles within each participant (such duplicates arise from $k$-nearest-neighbor interpolation between native participant surfaces and the fsaverage template space \citep{fischl1999high}, where distinct fsaverage vertices can inherit the response of the same native-surface vertex), dropping 646 voxels ($13\%$; $5{,}021 \to 4{,}375$).
Second, we then kept the top $50\%$ of voxels per (participant, fROI) combination, ranked by cross-validated NC-normalized predictivity ($r$) using the Matryoshka SAE features. Two participants had relatively few high-NCSNR voxels (P4: 117, P7: 113) and hence we did not further subsample these two participants' voxels. 
The final $2{,}296$ voxels are distributed across participants and fROIs as reported in Table~\ref{tab:generalization_voxels}.

% 1. Deduplication (5021 → 4375): Voxels with identical beta response patterns were removed per participant. This drops 646 voxels (13\%).
% 2. Regression subset (4375 → 2296): For six participants, exactly 50\% of deduplicated voxels were used. The two smallest participants (P4: 117, P7: 113) kept 100\%.

%  Strategy" Top 50\% per participant × fROI, keeping low-voxel participants (4,375 → 2,296): For each participant × fROI combination, the top 50\% of voxels by cross-validated NC Normalized R Fischer (Matryoshka featurizer) were selected. Participants with fewer than 200 total voxels (P4: 117, P7: 113) were exempt from cutting and retained all their voxels.

% All post the NCSNR filtering criterion. 

\begin{table}[ht]
\centering
\caption{Voxel counts for the fROI generalization analysis. Starting from the language fROI voxels (NCSNR $> 0.4$), the top 50\% by NC-normalized predictivity ($r$) were selected per participant $\times$ fROI, except for participants with fewer than 200 voxels (P4, P7), who were kept in full.}
\label{tab:generalization_voxels}
\begin{tabular}{lrrrrrrrrr}
\toprule
& \multicolumn{8}{c}{Participant} & \\
\cmidrule(lr){2-9}
fROI & P1 & P2 & P3 & P4 & P5 & P6 & P7 & P8 & Total \\
\midrule
IFGorb   & 52  & 39  & 39  & 0   & 16  & 24  & 0   & 21  & 191  \\
IFG      & 54  & 42  & 53  & 9   & 1   & 48  & 16  & 14  & 237  \\
MFG      & 64  & 98  & 25  & 7   & 59  & 11  & 16  & 59  & 339  \\
AntTemp  & 34  & 47  & 11  & 21  & 38  & 79  & 28  & 28  & 286  \\
PostTemp & 215 & 237 & 127 & 80  & 172 & 210 & 53  & 149 & 1243 \\
\midrule
Total    & 419 & 463 & 255 & 117 & 286 & 372 & 113 & 271 & 2296 \\
\bottomrule
\end{tabular}                                         
\end{table}

% Starting from deduplicated langfroi12345 voxels (NCSNR $> 0.4$), the top 50\% by NC-normalized predictivity ($r$) were selected per participant $\times$ fROI, except for participants with fewer than 200 voxels (P4, P7), who were kept in full.

\section{Language fROI Qualitative Feature Analysis}
\label{app:froiqualitative}

\begin{figure}
    \centering
    \includegraphics[width=\linewidth]{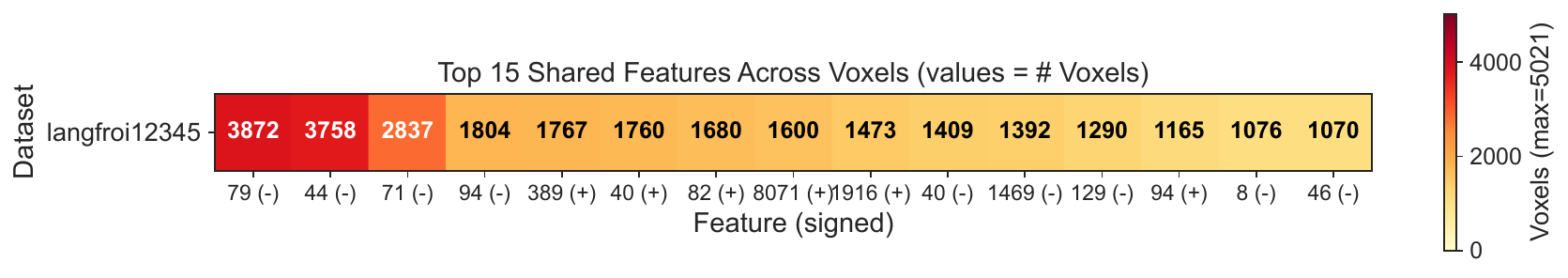}
    \caption{Signed Matryoshka feature prevalence across all participants and all language fROIs. We find many features from the most general Matryoshka bin, and many that arose when analyzing the voxel subtypes.}
    \label{fig:froiqualitative}
\end{figure}

We visualize the prevalence of signed Matryoshka features across all participants and fROIs in Fig.~\ref{fig:froiqualitative}. We first note that many of the most prevalent features are from the most general Matryoshka bin. Next, we note that many of these features appear in voxel subtype analysis in Section~\ref{Sec:Study1}. These features broadly reflect the interpretations of the PCs that have been identified in \citet{tuckute2025two}. Features -79, -94, +389, and +40 are all concordant with \textbf{Abstract} voxels, which are predominantly located in the language network. Features 44 and 71 are both driven more by sentences comprised of fewer tokens, and are thus suppressed by more complicated sentences. This reflects the prevalence of \textbf{Hard-to-Process} voxels in the language network.

\newpage
\section{Additional Hierarchical Representational Alignment Analyses}
\label{app:rep_align}
In this section, we present additional analyses on the utility of different Matryoshka features for predicting voxel responses.

\subsection{Extended fROI Analyses}

\begin{figure}
    \centering
    \includegraphics[width=.45\linewidth]{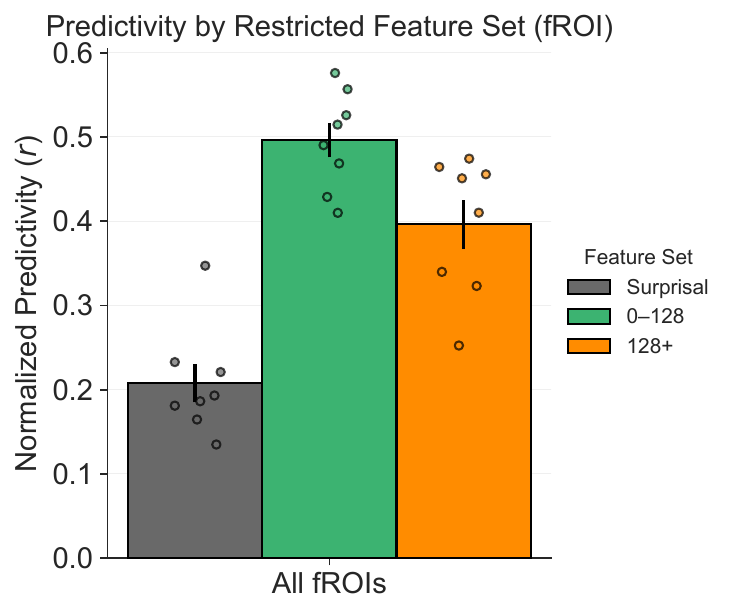}
    \caption{Predictivity of regressions restricted to using features from the most general Matryoshka bin vs. the union of all other features. We find that general features enable better predictivity.}
    \label{fig:fROIUnion}
\end{figure}

We present an extreme case of the analysis presented in Fig.~\ref{fig:granularity} --- we compare the predictivity of a regression restricted to features from the most general bin of Matryoshka features (128 features) to the predictivity of a regression restricted to the union of all \textit{other} feature bins ($>$ 30K features). In Fig.~\ref{fig:fROIUnion}, we find that the general feature bin enables better predictivity than the union of all other feature bins. This provides more support for the role of general features in predicting voxel tuning.

\subsection{Replication using Subtypes Data}

\begin{figure}
    \centering
    \includegraphics[width=\linewidth]{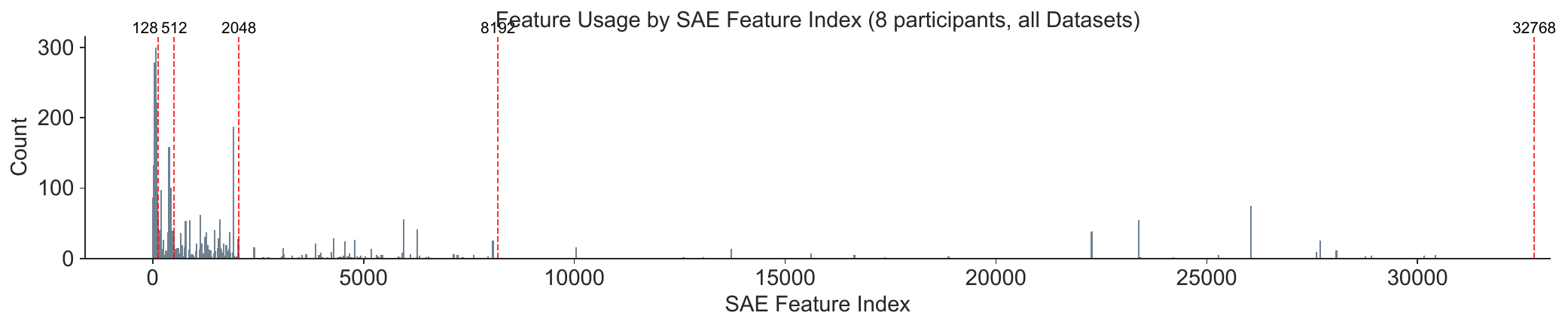}
    \caption{Histogram of how often each of the Matryoshka features is selected by encoding models in the voxel subtypes data.}
    \label{fig:CatHistogram}
\end{figure}

\begin{figure}
    \centering
    \includegraphics[width=\linewidth]{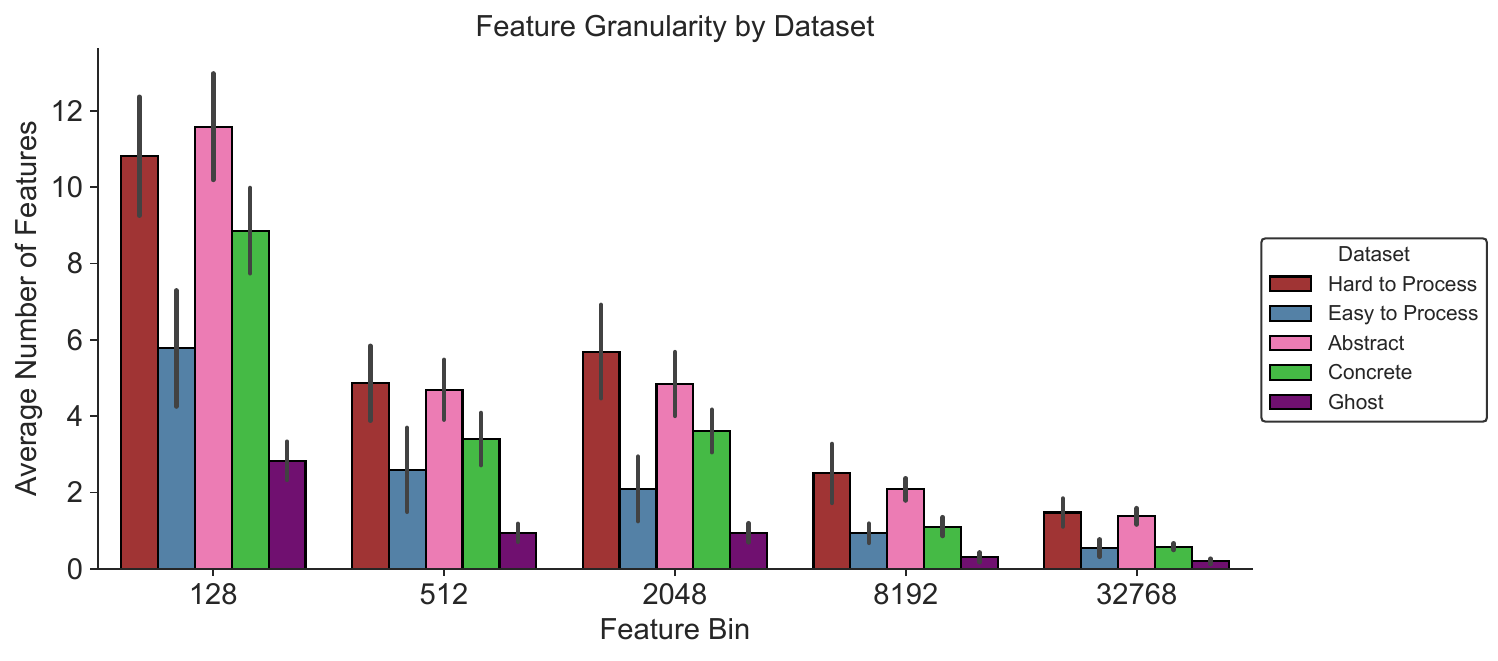}
    \caption{Average support set sizes (number of features selected by LASSO regression) per Matryoshka bin when predicting voxels in each subtype.}
    \label{fig:CatBarplot}
\end{figure}

\begin{figure}
    \centering
    \includegraphics[width=.5\linewidth]{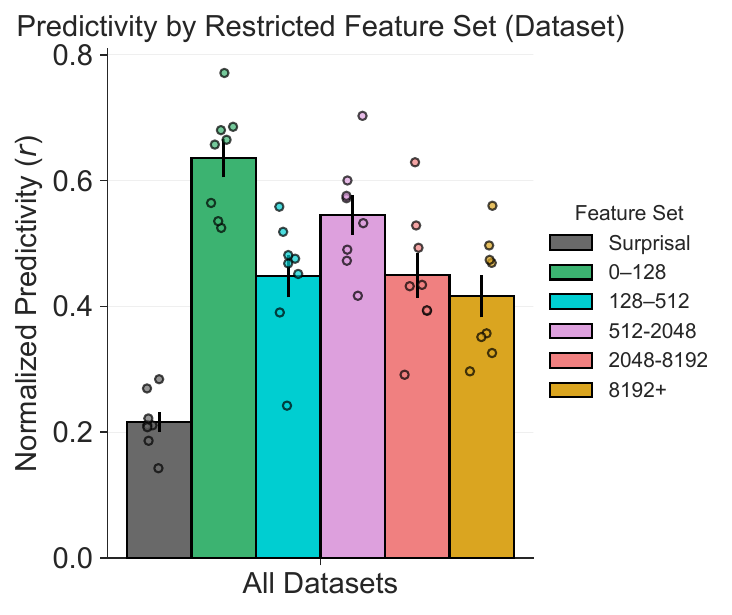}
    \caption{Performance of encoding models with feature sets restricted to individual Matryoshka feature bins.}
    \label{fig:CatBinPredictivity}
\end{figure}

\begin{figure}
    \centering
    \includegraphics[width=.5\linewidth]{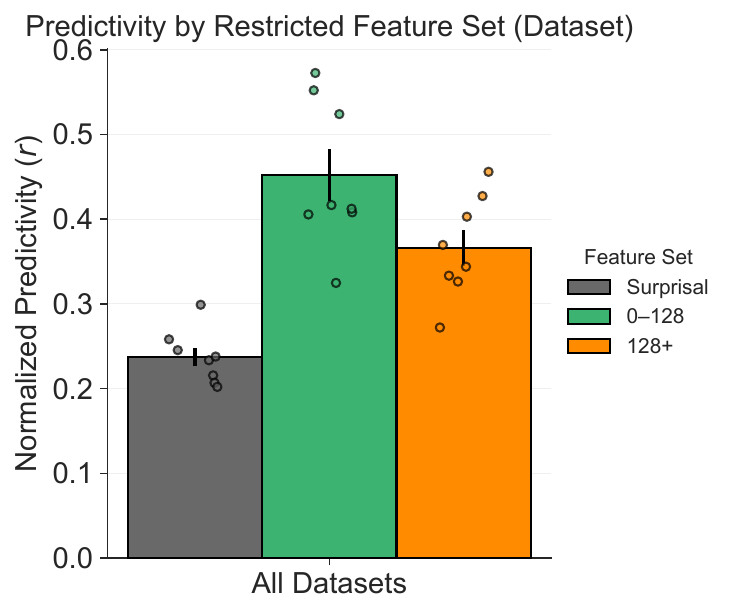}
    \caption{Predictivity of regressions restricted to using features from the most general Matryoshka bin vs. the union of all other features. We find that general features enable better predictivity for voxels in the PC-derived subtypes.}
    \label{fig:CatUnion}
\end{figure}

In this section, we replicate our feature granularity findings using the voxel subtypes data that were analyzed in Section~\ref{Sec:Study1}. Figs.~\ref{fig:CatHistogram}, \ref{fig:CatBarplot}, and \ref{fig:CatBinPredictivity} replicate the results presented in Fig.~\ref{fig:granularity}, and Fig.~\ref{fig:CatUnion} replicates the results in Fig.~\ref{fig:fROIUnion}.

\newpage
\section{Matryoshka SAE Feature Summary Statistics}\label{app:matryoshkastats}

\begin{figure}
\centering
\includegraphics[width=.99\linewidth]{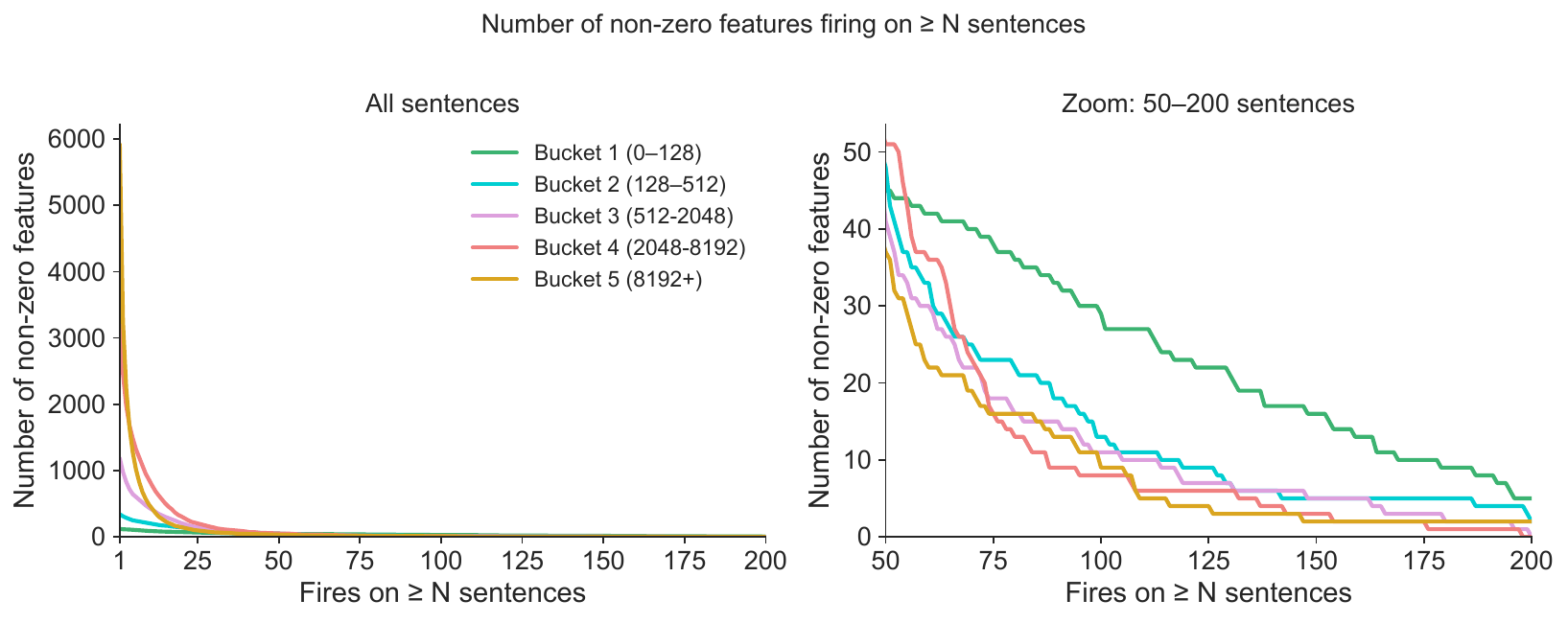}
\caption{Number of non-zero Matryoshka features firing on at least $N$ sentences, per bin. Left: full range, $N \in [1, 200]$. Right: zoom on $N \in [50, 200]$.}
\label{fig:matryoshka_bin_firing}
\end{figure}

In this section, we provide summary statistics for the Matryoshka SAE features (Gemma-2-2B, layer 12).
Table~\ref{tab:matryoshka_bin_stats} shows the count of non-zero features per Matryoshka bin on the 200-sentence stimulus set used in the main analyses, and Fig.~\ref{fig:matryoshka_bin_firing} shows the number of non-zero features firing on at least $N$ sentences as a function of $N$.

The table (\ref{tab:matryoshka_bin_stats}) shows that even in the fine-grained (late) bins there are still thousands of non-zero features (e.g., 3{,}397 in the 2048--8192 bin; 5{,}921 in 8192+), and at least one feature in every bin fires on nearly all 200 sentences (``Max fires'' of 197--200). 
The right panel of Fig.~\ref{fig:matryoshka_bin_firing} shows that around $N \approx 50$, all five bins converge to comparable non-zero feature counts (45, 48, 41, 51, 37 at $N{=}50$, respectively), even though the bins differ in raw size by orders of magnitude. 

Bin-level predictivity differences are therefore not a trivial consequence of e.g., all-zero features in the late bins.   

\begin{table}[]
\centering
\caption{Per-bin statistics for the Matryoshka SAE on the 200 sentences used in the main analyses encoding models.
``N non-zero'' is the count of the number of features that activate on at least one sentence; the parenthetical is the percentage of the total number of features in the bin. 
``Min / Max fires'' is the smallest / largest number of sentences (out of 200) any single non-zero feature in the bin fires on.}
\label{tab:matryoshka_bin_stats}
\begin{tabular}{clrrrr}
\toprule
Bin & Range & N feats & N non-zero & Min fires & Max fires \\
\midrule
1     & 0--128       & 128       & 122 (95.3\%)        & 1  & 200 \\
2     & 128--512     & 384       & 342 (89.1\%)        & 1  & 200 \\
3     & 512--2048    & 1{,}536   & 1{,}186 (77.2\%)    & 1  & 199 \\
4     & 2048--8192   & 6{,}144   & 3{,}397 (55.3\%)    & 1  & 197 \\
5     & 8192+        & 24{,}576  & 5{,}921 (24.1\%)    & 1  & 200 \\
\bottomrule
\end{tabular}
\end{table}

\end{document}